%% file: icml2024_conference.tex
\documentclass{article}

\usepackage{microtype}
\usepackage{graphicx}
\usepackage{subfigure}
\usepackage{booktabs} 

\usepackage{hyperref}

\usepackage[accepted]{icml2024}

\usepackage{amsmath}
\usepackage{amssymb}
\usepackage{mathtools}
\usepackage{amsthm}

\usepackage[capitalize,noabbrev]{cleveref}
\usepackage{multirow}
\usepackage{enumitem}

\usepackage{caption}
\usepackage{booktabs}
\usepackage{wrapfig}
\usepackage{tablefootnote}
\theoremstyle{plain}

\theoremstyle{definition}

\theoremstyle{remark}

\usepackage[textsize=tiny]{todonotes}

\newcommand{\method}{TS-LLM}
\newcommand{\revision}[1]{{#1}}

\icmltitlerunning{AlphaZero-Like Tree-Search can Guide Large Language Model Decoding and Training}

\begin{document}

\twocolumn[
\icmltitle{AlphaZero-Like Tree-Search can Guide\\ Large Language Model Decoding and Training}



\icmlsetsymbol{equal}{*}

\begin{icmlauthorlist}
\icmlauthor{Xidong Feng}{equal,ucl}
\icmlauthor{Ziyu Wan}{equal,sjtu}
\icmlauthor{Muning Wen}{sjtu}
\icmlauthor{Stephen Marcus McAleer}{cmu}
\\
\icmlauthor{Ying Wen}{sjtu}
\icmlauthor{Weinan Zhang}{sjtu}
\icmlauthor{Jun Wang}{ucl}
\end{icmlauthorlist}

\icmlaffiliation{ucl}{University College London}
\icmlaffiliation{sjtu}{Shanghai Jiao Tong University}
\icmlaffiliation{cmu}{Carnegie Mellon University}

\icmlcorrespondingauthor{Xidong Feng}{xidong.feng.20@ucl.ac.uk.}

\icmlkeywords{Machine Learning, ICML}

\vskip 0.3in
]



\printAffiliationsAndNotice{\icmlEqualContribution} 

\begin{abstract}

Recent works like Tree-of-Thought (ToT) and Reasoning via Planning (RAP) aim to augment the reasoning capabilities of LLMs by using tree-search algorithms to guide multi-step reasoning. These methods rely on prompting a pre-trained model to serve as a value function and focus on problems with low search depth. As a result, these methods will not work in domains where the pre-trained LLM does not have enough knowledge to serve as an effective value function or in domains that require long-horizon planning. To address these limitations, we present an AlphaZero-like tree-search learning framework for LLMs (termed TS-LLM), systematically illustrating how tree-search with a learned value function can guide LLM decoding. TS-LLM distinguishes itself in two key ways. (1) Leveraging a learned value function and AlphaZero-like algorithms, our approach can be generally adaptable to a wide range of tasks, language models of any size, and tasks of varying search depths. (2) Our approach can guide LLMs during both inference and training, iteratively improving the LLM. Empirical results across reasoning, planning, alignment, and decision-making tasks show that TS-LLM outperforms existing approaches and can handle trees with a depth of 64.
\end{abstract}

\input{sections/1-introduction}

\input{sections/2-related_work}
\input{sections/3-method}
\input{sections/4-experiment}
\input{sections/5-conclusion}

\newpage
\section{Impact Statements}
This paper presents work whose goal is to advance the field of Deep Learning. There are potential societal consequences of our work, none which we feel must be specifically highlighted here.
\nocite{langley00}

\bibliography{iclr2024_conference}
\bibliographystyle{icml2024}

\newpage
\appendix
\onecolumn

\input{sections/appendix}


\end{document}

%% file: sections/1-introduction.tex
\section{Introduction}

Large language models (LLMs) \citep{OpenAI2023GPT4TR, touvron2023llama} have demonstrated their potential in a wide range of natural language tasks. A plethora of recent studies have concentrated on improving LLMs task-solving capability, including curation of larger and higher-quality general or domain-specific data \citep{touvron2023llama, zhou2023lima, gunasekar2023textbooks, feng2023chessgpt, taylor2022galactica}, more sophisticated prompt design \citep{wei2022cot0, zhou2022l2mcot, creswell2022sicot}, or better training algorithms with Supervised Learning or Reinforcement Learning (RL) 
\citep{dong2023raft, gulcehre2023reinforced, rafailov2023direct}. When training LLMs with RL, LLMs' generation can be naturally formulated as a Markov Decision Process (MDP) and optimized with specific objectives. Following this formulation, ChatGPT \citep{ouyang2022training} emerges as a notable success, optimizing LLMs to align human preference by leveraging RL from Human Feedback (RLHF) \citep{christiano2017deep}.

LLMs can be further guided with planning algorithms such as \textbf{tree search}. 
Preliminary work in this field includes Tree-of-Thought (ToT) \citep{yao2023tot,long2023large} with depth/breadth-first search and Reasoning-via-Planing (RAP) \citep{hao2023rap} with MCTS. 
They successfully demonstrated a performance boost of searching on trees expanded by LLM through self-evaluation. 
\begin{figure*}[t]
\vspace{-5pt}
    \centering
    \includegraphics[width=0.9\linewidth]{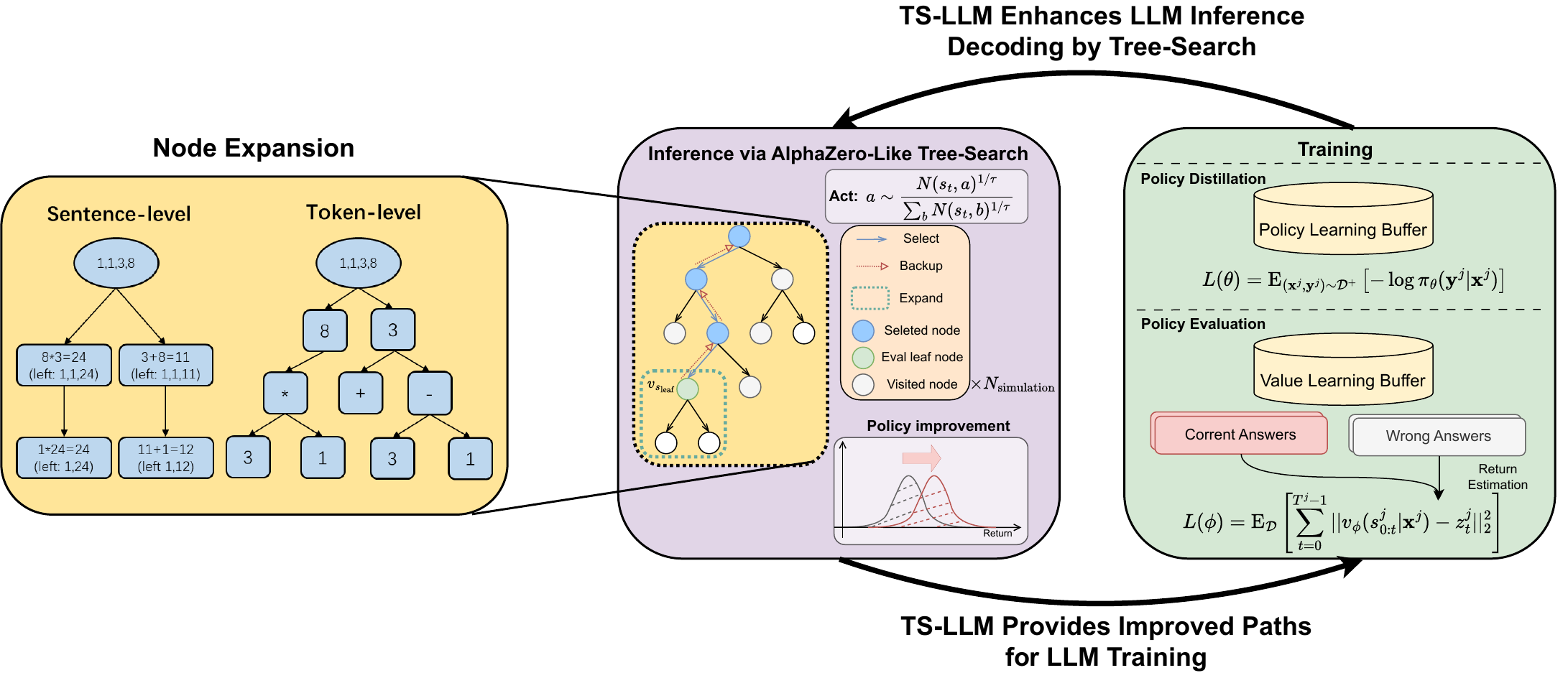}
    \caption{(a) Left: Two node expansion paradigms on Game24: sentence-level and token-level. We adopt sentence-level setting in this task.
    (b) Right: \method~consists of an iterative process over tree-search and training. First, TS-LLM enhances LLM inference by tree-search to obtain improved trajectories, augmenting the training set. LLM can be further trained to improve by conducting policy distillation and value function learning (policy evaluation) over the augmented training set.}
    \label{fig:main_fig}
    \vspace{-15pt}
\end{figure*}
Despite these advances, current methods come with distinct limitations. First, 
the value functions in the tree-search algorithms are obtained by prompting LLMs. As a result, such algorithms lack general applicability and heavily rely on both well-designed prompts and the robust capabilities of advanced LLMs. Beyond the model requirements, we will also show in Sec. \ref{exp:learned_value} that such prompt-based self-evaluation is not always reliable.
Second, ToT and RAP use BFS/DFS and MCTS for tree search, restricting their capabilities to relatively simple and shallow tasks. They are capped at a maximum depth of only 10 or 7, which is significantly less than the depth achieved by AlphaZero in chess or Go~\cite{silver2017alphazero}. As a result, ToT and RAP might struggle with complex problems that demand large analytical depths and longer-term planning horizons, decreasing their scalability.
To address these problems, we introduce tree-search enhanced LLM (TS-LLM), an AlphaZero-like framework that utilizes tree-search to improve LLMs' performance on general natural language tasks. TS-LLM extends previous work to AlphaZero-like deep tree-search with a learned LLM-based value function which can guide the LLM during both inference and training. Compared with previous work, TS-LLM has the following two new features:
\begin{itemize}[leftmargin=*]
    \item\textbf{\method~offers a generally applicable and scalable pipeline.} It is generally \textbf{applicable}: With a learned value function, TS-LLM can be applied to various tasks and LLMs of any size. Our learned value function can be more reliable than the prompt-based counterpart and does not require any well-designed prompts or advanced, large-scale LLMs. Our experiments show that TS-LLM can work for LLMs ranging from 125M to 7B parameters, providing better evaluation even compared with GPT-3.5. TS-LLM is also \textbf{scalable}: TS-LLM can conduct deep tree search, extending tree-search for LLM generation up to a depth of 64. This is far beyond 10 in ToT and 7 in RAP.
    \item \textbf{\method~can potentially serve as a new LLM training paradigm beyond inference decoding.} By treating the tree-search operation as a policy improvement operator, we can conduct an iterative processes of improving the policy via tree search and then improving the policy through distillation and the value function through the ground-truth training labels on the tree search trajectories.
\end{itemize}
Through comprehensive empirical evaluations on reasoning, planning, alignment, and decision-making tasks, we present an in-depth analysis of the core design elements in \method, delving into the features, advantages, and limitations of different variations. This showcases TS-LLM's potential as a universal framework to guide LLM decoding and training.

%% file: sections/2-related_work.tex
\vspace{-5pt}
\section{Related Work}
\label{sec:related_work}
\textbf{Multistep Reasoning in LLMs}
Multistep reasoning in language models has been widely studied, from improving the base model \citep{chung2022flan, fu2023specializing, lewkowycz2022minerva} to prompting LLMs step by step \citep{kojima2022zscot,wei2022cot0, wang2022cot-sc, zhou2022l2mcot}. Besides, a more relevant series of work has focused on enhancing reasoning through evaluations, including learned reward models \citep{uesato2022dm-orm, lightman2023prm800k} and self-evaluation \citep{shinn2023reflexion, madaan2023selfrefine}. 
In this work, we apply evaluations to multistep reasoning tasks and token-level RLHF alignment tasks by learning a value function and reward model under the setting of a multistep decision-making process.

\textbf{Search-Guided Reasoning in LLMs}
While most CoT approaches have used a linear reasoning structure, recent efforts have been made to investigate non-linear reasoning structures such as trees \citep{jung2022maieutic, zhu2022mctscooperative}. 
More recently, various approaches for searching on trees have been applied to find better reasoning paths, e.g. beam search in \citet{xie2023decomposition}, depth-/breadth-first search in \citet{yao2023tot} and Monte-Carlo Tree Search in \citep{hao2023rap}.
Compared with these methods, \method~is a tree search guided LLM decoding and training framework with a learned value function, which is more generally applicable to both reasoning tasks and other scenarios like RLHF alignment tasks. 
Due to the limit of space, we leave a comprehensive comparison in Appendix~\ref{app:comparsion-related_work}.
 
\textbf{Finetuning LLMs with Augmentation} 
Recent efforts have also been made to improve LLMs with augmented data. Rejection sampling is a simple and effective approach for finetuning data augmentation to improve LLMs' ability on single/multiple task(s) such as multistep reasoning \citep{yuan2023rft, zelikman2022star} and alignment with human preference  \citep{dong2023raft, bai2022constitutional}. 
Given an augmented dataset, reinforcement learning approaches have also been used to finetune LLMs \citep{gulcehre2023reinforced, luo2023wizardmath}.
Compared to previous works, \method~leverages tree search as a policy improvement operator to generate augmented samples to train both the LLMs and the value function.

%% file: sections/3-method.tex
\section{Enhancing LLMs with Tree Search}









\label{sec:enhancing}
In this section, we propose a versatile tree-search framework \method~for guiding LLMs decoding and training. We conduct a systematic and comprehensive analysis of its key components. \method~is summarized in Fig.~\ref{fig:main_fig}.


\vspace{-5pt}
\subsection{Problem Formulation}
\label{sec:method-problem_formulation}
We formulate the language generation process as a multi-step Markov Decision Process (MDP). The particularity of natural language tasks is that both the action and state are in language space. LLMs can serve as a policy $\pi_\theta$ that samples sequences of tokens as actions. Assuming the length of the output sequence and input prompt are $T$ and $L$ respectively, the probability for an LLM policy to produce an output sequence $\mathbf{y}=(y_0, y_1, \ldots, y_{T-1})$ conditioned on a prompt (input prefix) $\mathbf{x}=(x_0, x_1, \ldots, x_{L-1})$ is:
$
    \pi_\theta(\mathbf{y} | \mathbf{x}) = \prod_{t=0}^{T-1} \pi_\theta(y_{t}| \mathbf{x}_{0:L-1}, \mathbf{y}_{0:t-1}).
$

For a given natural language task, we can define a reward function $R(y_t|\mathbf{x}_{0:L-1}, \mathbf{y}_{0:t-1})$ as the task performance feedback for intermediate generation $y_t$ at timestep $t$. Due to the lack of large-scale and high-quality intermediate reward labels for general tasks, it is usually a sparse reward setting where any intermediate reward from 
the first $T-1$ timestep is zero except the last $T$-th step. A typical case can be RLHF alignment task, where LLM can receive the reward signal after it completes the full generation. 
Following the same logic, $\mathbf{y}$ can also be viewed as a sequence of sentences.

Given the problem formulation above, we successfully transfer the problem of better generation to optimization for higher cumulative reward. In this paper, we focus on how we can optimize it with \textbf{tree-search algorithms}. A specific natural language task typically predefines the state space (with language) and reward function (with task objective/metrics). What remains is the definition of action space, or in the context of tree-search algorithm, the action node.

Tree search algorithms have validated their effectiveness for different action spaces in traditional RL research, including discrete action space \citep{silver2017alphazero, schrittwieser2020mastering} and continuous action space \citep{hubert2021learning}. For tree-search on LLMs, we consider the following two action space designs, as shown in the left side of Fig.~\ref{fig:main_fig}.

\textbf{Sentence-level action nodes}: For the tasks that have a step/sentence-level structure(e.g. chain-of-thought reasoning), it is natural to treat each thought as a sentence-level action node. This is also the technique adopted by ToT \citep{yao2023tot} and RAP \citep{hao2023rap}. For each non-terminal node, the search tree is expanded by sampling several possible subsequent intermediate steps and dropping the duplicated generations.

\textbf{Token-level action nodes}: Analogous to tree-search in discrete action space MDP, we can treat each token as a discrete action for LLM policy and the tree search can be conducted in token-level. For those tasks in which intermediate steps are not explicitly defined(e.g. RLHF), splitting an output sequence into tokens might be a good choice.

\revision{Typically, the search space is determined by two algorithm-agnostic parameters, the \textbf{tree max width $w$} and \textbf{tree max depth $d$}. In LLM generation, }both action space designs have their advantages and limitations \revision{over the search space}. By splitting the generation into sentences, \textbf{sentence-level action nodes} provide a relatively shallow tree (\revision{low tree max-depth}), simplifying the tree-search process. However, the large sample space of sentence-level generation makes full enumeration of all possible sentences infeasible. We have to \revision{set a maximum tree width $w$} to subsample $w$ nodes during the expansion, similar to the idea of Sampled MuZero~\citep{hubert2021learning} (The node will be fixed once it is expanded). Such subsampling results in the gap, determined by $w$, between the tree-search space and the LLM generation space. For \textbf{token-level action nodes}, though it can get rid of the search space discrepancy and extra computational burdens, it greatly increases the depth of the tree, making tree-search more challenging.


\subsection{Guiding LLM Inference Decoding with Tree Search}
\label{sec:3.3}
One of the benefits of tree-search algorithms is that they can optimize the cumulative reward by mere search, without any gradient calculation or update. In this section, given a fixed LLM policy, we present the full pipeline to illustrate how to guide LLM inference decoding with tree search approaches.
\subsubsection{Learning an \revision{LLM-based} Value Function}
\label{sec-3.2.1-learn_value_function}
For tree-search algorithms, how to construct reliable value function $v$ and reward model $\hat{r}$ is the main issue. ToT and RAP obtain these two models by prompting advanced LLMs, such as GPT-4 or LLaMA-33B. To make the tree search algorithm generally applicable, our method leverages a learned \revision{LLM-based} value function $v_\phi(s)$ conditioned on state $s$ and a learned final-step outcome reward model (ORM) $\hat{r}_\phi$ since most tasks can be formulated as sparse-reward problems~\citep{uesato2022dm-orm}. 
Since we mainly deal with language-based tasks, we utilize a shared value network and reward model whose structure is a decoder-only transformer with an MLP to output a scalar on each position of the input tokens. \revision{And typically, LLM value's decoder is adapted from original LLM policy $\pi_{\theta}$'s decoder, or the LLM value $v_\phi$ and policy $\pi_{\theta}$ can have a shared decoder\citep{silver2017alphazero}.} For a sentence-level expanded intermediate step $s_t$, we use the prediction scalar at the last token as its value prediction $v_\phi(s_t)$. The final reward can be obtained at the last token when feeding the full sentences $(\mathbf{x}_{0:L-1}, \mathbf{y}_{0:T-1})$ into the model.

Therefore, we use language model $\pi_\theta$ as the policy to sample generations using the task training dataset. With true label or a given reward function in training data, a set of sampled tuple $\mathcal{D}_\text{train} = \{ (\mathbf{x}^j, \mathbf{y}^j, r^j) \}_j$ of size $|\mathcal{D}_\text{train}|$ can be obtained, where $\mathbf{x}^j$ is the input text, $\mathbf{y}^j=s^j_{0:T^j-1}$ is the output text of $T^j$ steps and $r^j=R(\mathbf{y}^j | \mathbf{x}^j)$ is the ground-truth reward. Similar to the critic training in most RL algorithms, we construct the value target $z^j_t$ by TD-$\lambda$ \citep{sutton1988learning} or MC estimate \citep{sutton2018reinforcement} on each single step $t$ . 
The value network is optimized by mean squared error:
\begin{equation}
\label{equation:value_loss}
    L(\phi) = \displaystyle \text{E}_{\mathcal{D}}\left[ \sum\limits_{t=0}^{T^j-1}||v_{\phi}(s^j_{0:t}|\mathbf{x}^j) - z^j_t ||_2^2 \right].
\end{equation}
The ORM $\hat{r}_\phi(\mathbf{y}_{0:T-1} | \mathbf{x}_{0:L-1})$ is learned with the same objective. Training an accurate value function and ORM is quite crucial for the tree-search process as they provide the main guidance. We will further illustrate how to learn a reliable value function and ORM in our experiment section.

\subsubsection{Tree Search Algorithms} 
\label{sec:tree search algorithms}

Given a learned value function, in this section, we present five types of tree-search algorithms. We leave detailed background, preliminaries, and comparisons in Appendix \ref{apx:ts_algos}.

\textbf{Breadth-First and Depth-First Search \revision{With Value Function-Based Tree-Pruning(BFS-V/DFS-V)}}: These two search algorithms were adopted in ToT \citep{yao2023tot}. The core idea is to utilize the value function to prune the tree for efficient search, while such pruning happens in tree breadth or depth respectively. BFS-V can be regarded as a beam-search with cumulative reward as the objective.

\textbf{MCTS}: This approach was adopted in RAP \citep{hao2023rap}, which refers to classic MCTS\citep{kocsis2006mcts-bandit}. It back-propagates the value on the terminal nodes, relying on a Monte-Carlo estimate of value, and it starts searching from the initial state node. 

In addition to these algorithms adopted by ToT and RAP, we consider two new variants of AlphaZero-like tree-search.

\textbf{MCTS with Value Function Approximation} (named as MCTS-$\alpha$): This is the MCTS variants utilized in AlphaZero \citep{silver2017alphazero}. Starting from the initial state, we choose the node of state $s_t$ as the root node and do several times of search simulations consisting of \textit{select, expand and evaluate} and \textit{backup}, where the leaf node value evaluated by the learned value function will be backpropagated to all its ancestor nodes. After the search, we choose an action proportional to the root node's exponentiated visit count, i.e. $ a \sim \frac{N(s_t, a)^{1 / \tau}}{\sum_b N(s_t, b)^{1 / \tau}}$, and move to the corresponding next state. The above iteration will be repeated until finished. MCTS-$\alpha$ has two main features. Firstly, MCTS-$\alpha$ cannot trace back to its previous states once it takes an action. So it cannot restart the search from the initial state unless multiple searches are conducted which will be discussed in Section \ref{sec-3.2.3-multi_search_aggregation}. Secondly, in contrast to MCTS, MCTS-$\alpha$ utilizes a value function so it can conduct the backward operation during the intermediate steps, without the need to complete the full generation to obtain a Monte-Carlo estimate.

\textbf{MCTS-Rollout}: Combining the features from MCTS and MCTS-$\alpha$, we propose a new variant MCTS-Rollout for tree search. Similar to MCTS, MCTS-Rollout always starts from the initial state node. It further does the search simulations analogous to MCTS-$\alpha$, and the \textit{backup} process can happen in the intermediate step with value function. It repeats the operations above until the process finds $N$ complete answers or reaches the computation limit (e.g. maximum number of tokens.) \revision{MCTS-Rollout can be seen as an offline version of MCTS-$\alpha$ so they may have similar application scope. The only difference is that MCTS-Rollout can scale up the token consumption for better performance since it always reconducts the search from the beginning.}

\begin{table*}[t]
    \centering
    \caption{Task setups. \revision{The node, tree max with and tree max depth are search space parameters. Refer to Appendix \ref{apx: node_expansion} and \ref{apx:hyper_selection} for how max tree-width and tree-depth are determined.}}
    \label{tab:exp-task_setup}

    \small
    \begin{tabular}{c c c c c c}
        \toprule
        & & &\multicolumn{3}{c}{\revision{Search Space Hyperparameters}} \\
\cmidrule{4-6} 
        Task & Category  & Train/test size & Node & Tree Max width & Tree Max depth\\
        \midrule
        GSM8k    & Mathematical Reasoning  & 7.5k / 1.3k & Sentence & 6  & 8   \\
        Game24   & Mathematical Planning & 1.0k / 0.3k& Sentence  & 20 & 4 \\
        PrOntoQA & Logical Reasoning & 4.5k / 0.5k & Sentence & 6  & 15\\
        RLHF     & Alignment    & 30k / 3k     & Token & 50 & 64 \\
        Chess Endgame & Decision Making & 0.1M/0.6k & Sentence & 5 & 50 \\
        \bottomrule
        \vspace{-20pt}
    \end{tabular}

\end{table*}

\subsubsection{Multiple Search and Search Aggregation}
\label{sec-3.2.3-multi_search_aggregation}
Inspired by \citet{wang2022cot-sc} and \citet{uesato2022dm-orm} that LLM can improve its performance on reasoning tasks by sampling multiple times and aggregating the candidates, TS-LLM also has the potential to aggregate \revision{$N$} complete answers generated by multiple tree searches or multiple generations from one search (set BFS beam size $>1$).

\revision{When conducting multiple tree searches, we usually adopt \textbf{intra-tree search} setting.} Intra-tree search conducts multiple tree searches on the same tree, \revision{thus the state space is exactly the same}. Such a method is computationally efficient as the search tree can be reused multiple times. However, the diversity of multiple generations might decrease because the former tree search might influence the latter tree searches. Also, the search space is limited in sentence-level action space because they will be fixed once expanded across multiple tree searches.

\revision{We refer to Appendix~\ref{appendix:details-of-aggregation} for an alternative setting called Inter-tree Search where we allow resampling in the expansion process, and without further specification, all settings in our paper are under the intra-tree search setting.} Our next step is to aggregate these search results to obtain the final answer. With a learned ORM, we consider the following three different aggregation methods:

\textbf{Majority-Vote} \citet{wang2022cot-sc} aggregates answers using majority vote: $f^*=\arg\max_{f} \sum\nolimits_{\mathbf{y}^j} \mathbf{1}_{\text{final\_ans}(\mathbf{y}^j)=f}$, where $\mathbf{1}$ is the indicator function.
\\
\textbf{ORM-Max.} Given an outcome reward model, the aggregation can choose the answer $f$ with maximum final reward, $f^*=\text{final\_ans}(\arg\max_{\mathbf{y}^j} \hat{r}_\phi(\mathbf{y}^j | \mathbf{x}^j))$.
\\
\textbf{ORM-Vote.} Given an outcome reward model, the aggregation can choose the answer $f$ with the sum of reward, namely $f^*=\arg\max_{f} \sum\nolimits_{\mathbf{y}^j;\text{final\_ans}(\mathbf{y}^j)=f}\hat{r}_\phi(\mathbf{y}^j | \mathbf{x}^j)$.

\subsection{Enhancing LLM Training with Tree Search}
\label{sec-3.3-trian_with_TS}
In section \ref{sec:3.3} we discuss how tree-search can guide LLM's decoding process during inference time. Such guidance leads to a better decoding strategy and improves the performance of given tasks. In other words, tree-search guidance can serve as a policy improvement operator. Based on this, we propose a new training and finetuning paradigm.

Assume we have an initial LLM policy $\pi_{\theta_\text{old}}$ (trained by conducting supervised finetuning over the original training set) and initial LLM value and ORM: $v_{\phi_\text{old}}$, $\hat{r}_{\phi_\text{old}}$ (trained by Equ.~\ref{equation:value_loss} from sampling the original training questions), we can have the following iterative process:

\textbf{Policy Improvement}: We conduct tree-search over training set based on $\pi_{\theta_\text{old}}$, $v_{\phi_\text{old}}$, and $\hat{r}_{\phi_\text{old}}$ to obtain improved generations, resulting in the augmented dataset $\mathcal{D}$ and also the filtered positive examples $\mathcal{D}^+$.

\textbf{Policy Distillation:} With the tree-search-improved dataset $\mathcal{D}^+$, by imitating the tree-search positive trajectories, LLM policy can be further improved to $\pi_{\theta_\text{new}}$ with supervised loss.
\begin{equation}
    L(\theta)
    =\text{E}_{(\mathbf{x}^j, \mathbf{y}^j) \sim \mathcal{D}^+}
    \left[
        -\log \pi_{\theta}(\mathbf{y}^j | \mathbf{x}^j)
    \right].
\end{equation}
\textbf{Policy Evaluation}: We train value function $v_{\phi_\text{new}}$ and ORM $\hat{r}_{\phi_\text{new}}$ over the augumented dataset $\mathcal{D}$  under loss of Equ~\ref{equation:value_loss}.

These three processes can be conducted cyclically to iteratively refine the LLM. Such iterative process belongs to generalized policy iteration \citep{sutton2018reinforcement}, which is also the procedure used in AlphaZero's training. In our case, the training process involves finetuning three networks on the tree-search augmented dataset: (1) Policy network $\pi_\theta$: Use cross-entropy loss with trajectories' tokens as target (2) Value network $v_\phi$: Mean squared error loss with trajectories' temporal difference (TD) or Monte-Carlo (MC) based value estimation as target, and (3) ORM $\hat{r_\phi}$: Mean squared error loss with trajectories' final reward as target.

\subsection{Tree Search's Extra Computation Burdens}
\label{sec: ts_compute}
Tree-search algorithms will inevitably bring in additional computation burdens, especially in the node expansion phase for calculating legal child nodes and their corresponding value. Prior methodologies, such as ToT and RAP, tend to benchmark their performance against baseline algorithms using an equivalent number of generation paths (named Path@$N$). This approach overlooks the additional computational demands of the tree-search process. We also refer readers to Appendix~\ref{appendix:wall-time-and-engineering} for discussion about the computation efficiency and engineering challenges.

A more fair comparison requires monitoring the number of tokens generated for node expansion. This provides a reasonable comparison of algorithms' performance when operating under comparable token generation conditions. We address this issue in our experiments (Sec. \ref{exp:algos}).

%% file: sections/4-experiment.tex
\vspace{-5pt}
\section{Experiments}
In this section, we conduct thorough experiments to address and analyze each subsection we mention in Section~\ref{sec:enhancing}.




\subsection{Experiment Setups}
\label{sec:es}

\textbf{Task Setups} For a given MDP, the nature of the search space is primarily characterized by two dimensions: depth and width. To showcase the efficacy of tree-search algorithms across varied search spaces, we evaluate all algorithms on five tasks with different search widths and depths, including the mathematical reasoning task
GSM8k \citep{cobbe2021gsm8k}, mathematical planning task Game24 \citep{yao2023tot}, logical reasoning task PrOntoQA \citep{saparov2022prontoqa}, RLHF alignment task using synthetic RLHF data \citep{rlhf_data}, and chess endgames \citep{abdulhai2023lmrl}. The specific task statistics and search space hyperparameters are listed in Table \ref{tab:exp-task_setup}. These hyperparameters, especially max search width and search depth, are determined by our exploration experiments. They can effectively present the characteristics of a task \revision{and define its} search space. Refer to Appendix~\ref{appendix:task_setup} for more details of our evaluation tasks.

\textbf{Benchmark Algorithms.} We compare ToT-GPT3.5 and TS-LLM in Table \ref{tab:llm_value_function} to verify the effectiveness of learned value function. All tree-search algorithms will be benchmarked, including MCTS-$\alpha$, MCTS-Rollout, MCTS, BFS-V, and DFS-V. Note that BFS-V, DFS-V, and MCTS are TS-LLM's variants instead of ToT\citep{yao2023tot} or RAP\citep{hao2023rap} baselines because we adopt a learned value function rather than prompting LLM. 
We compare these variants with direct decoding baselines, including CoT greedy decoding, and CoT with self-consistency \revision{\citep{wang2022cot-sc}} (denoted as CoT-SC). Considering the search space gap between direct decoding and tree decoding (especially the sentence-level action node), we include the CoT-SC-Tree baseline which conducts CoT-SC over the tree's sentence nodes. 

\textbf{Model and Training Details.}
    For the rollout policy used in tree-search, we use LLaMA2-7B \citep{touvron2023llama2} on three reasoning tasks, and GPT-2-small (125M) on the RLHF task and Chess endgame. All LLMs will be first supervise-finetuned \revision{(SFT)} on the training set, enabling their zero-shot CoT ability. For value and ORM training, the data are generated by sampling the SFT policy's rollouts on the training set. \revision{Our policy LLM and value LLM are two separate models but are adapted from the same base model. Refer to Appendix \ref{apx:shared_decoder} for experiments on shared models.}
\vspace{-10pt}
\subsection{Results and Discussions}
\subsubsection{Learned Value function (Sec~\ref{sec-3.2.1-learn_value_function})} 
\label{exp:learned_value}
\begin{table}[t]
\small
    \centering
    \caption{ToT-BFS Path@1 results with different combinations of policy and value. LLaMA-SFT and LLaMA-V refer to the trained policy and value, LLaMA and GPT-3.5 (ToT) refer to the prompt-based model for policy or value. LLaMA-V dominates compared with prompt-based value.}
    \label{tab:llm_value_function}
    \begin{tabular}{cccc}
        \toprule
        Task & Policy & Value & Accuracy(\%) \\
        \midrule
        \multirow{5}{*}{GSM8K} & GPT-3.5 & GPT-3.5 (ToT) & 72.7 \\
          & GPT-3.5 & LLaMA-V (Ours) & \textbf{74.0}\\
          \cmidrule{2-4}
         & LLaMA-SFT & LLaMA (ToT) & 37.4 \\
         & LLaMA-SFT & GPT-3.5 (ToT) & 45.8 \\
         & LLaMA-SFT & LLaMA-V (Ours) & \textbf{52.5}\\
         \midrule
        \multirow{5}{*}{Game24}& GPT-3.5 & GPT-3.5 (ToT) & 15.5 \\
          & GPT-3.5 & LLaMA-V (Ours) & \textbf{19.1}\\
          \cmidrule{2-4}
         & LLaMA-SFT & LLaMA (ToT) & 9.2\\
         & LLaMA-SFT & GPT-3.5 (ToT) & 21.0 \\
         & LLaMA-SFT & LLaMA-V (Ours) & \textbf{64.8}\\
        \bottomrule
    \end{tabular}
    \vspace{-15pt}
\end{table}
We successfully show that \textbf{the learned value function can be more reliable than prompt-based GPT-3.5}, even though GPT-3.5 is much stronger than LLaMA2-7B. In Table \ref{tab:llm_value_function}, we conduct BFS Path@1 comparisons with different combinations of policy and value over Game 24 and GSM8K. The policy choices include few-shot GPT-3.5 and our supervised finetuned LLaMA2-7B. For the value, we utilize prompt-based GPT-3.5/LLaMA2-7B (TOT) and our learned value function LLaMA2-V. Even though the few-shot GPT-3.5 policy is an out-of-distribution policy for LLaMA2-V's evaluation, LLaMA2-V still presents dominant performance over prompt-based GPT-3.5/LLaMA2-7B in all settings. The phenomenon of LLM’s limited self-evaluation ability by prompting aligns with other papers \citep{huang2023large, stechly2023gpt}. This finding largely increases the necessity of a learned value function.

\begin{table*}[t]
    \centering
    \small
    \caption{Path@1 and Equal-token results of all TS-LLM variants and the CoT baseline with \#token. For the alignment task and Chess Endgame, SC$_\text{ORM}$ means the best score of the sampled candidates, and SC$_\text{MAJ}$ refers to the average. We only present BFS-V in Path@1 because MCTS/BFS-V/DFS-V degenerates to greedy value search. For the Equal-token setting, We do not show the results of BFS-/DFS-V/MCTS$_\text{ORM}$ in GSM8K since token consumption is similar in the Path@1 setting.
    } 
    \label{tab:path1_result}
    \vspace{-3mm}
    \begin{tabular}{cc cc cc cc cc cc cc}
        \toprule
         \multirow{2}{*}{Setting}&\multirow{2}{*}{Method} & \multicolumn{6}{c}{Performance(\%) / \# Tokens} & \multicolumn{2}{c}{Reward / \# Forward} & \multicolumn{2}{c}{Win Rate / \# Tokens}\\
         \cmidrule{3-12}
           & & \multicolumn{2}{c}{GSM8k} & \multicolumn{2}{c}{Game24} & \multicolumn{2}{c}{PrOntoQA} & \multicolumn{2}{c}{RLHF(token-level)} & \multicolumn{2}{c}{Chess Endgame} \\
         \midrule
         \multirow{4}{*}{Path@1}& CoT-greedy & 41.4 & 98 & 12.7 & 76 & 48.8 & 92 & 0.318 & 57.8 & 58.14 & 37.4 \\
         & BFS-V (Ours)  & \textbf{52.5} & 485 & 64.8 & 369 & 94.4 & 126 & -1.295 & 61.8 & 67.75 & 402 \\
         
         & MCTS-$\alpha$ (Ours)   & 51.9 & 561 & 63.3 & 412 & \textbf{99.4}& 190 & \textbf{2.221}  & 186   & 96.90 & 797\\
         
         & MCTS-Rollout (Ours)  & 47.8 & 3.4k & \textbf{71.3} & 670 & 96.9 & 210 & 1.925 & 809 & \textbf{98.76} & 615  \\
         
         
         \midrule

         \multirow{5}{*}{Equal-Token} & CoT-SC$_\text{MAJ}$  & 46.8  & 500 & 14.6  & 684 & 61.1  & 273  & -0.253  & 580 & 9.84 & 782 \\
         
         &CoT-SC$_\text{ORM}$  & \textbf{52.3}  & 500 & 50.6  & 684 & 83.2  & 273  & 1.517  & 580 & 73.80 & 782\\

         

         \cmidrule{2-12}
         & BFS-V$_\text{ORM}$ (Ours)  & - & - & 70.90 & 1.6k & - & - & -1.065 & 613 & 93.18 & 854 \\
         & DFS-V$_\text{ORM}$ (Ours)  & - & - & 69.09 & 962  & 96.4  & 195 & -0.860 & 86 & 71.01 & 511 \\
         & MCTS$_\text{ORM}$ (Ours)   & - & - & 69.34 & 649  & \textbf{99.6}  & 182 & 0.160 & 592 & 94.26 & 706 \\
         \bottomrule
    \end{tabular}
    \vspace{-3mm}
\end{table*}

\subsubsection{Performance on different algorithms (Sec~\ref{sec:tree search algorithms}, Sec~\ref{sec: ts_compute})} 
\label{section:exp-results}
    

\label{exp:algos}
With a reliable learned value function, we compare the performance of different generation methods. First, in the upper part of Table~\ref{tab:path1_result}, we present the Path@1 results of MCTS-$\alpha$ and MCTS-Rollout compared to BFS-V (BFS-/DFS-V and MCTS degenerate to greedy value tree-search in path@1 case) and CoT-Greedy. The experiment results show that \textbf{AlphaZero-like search algorithms, MCTS-$\alpha$ and MCTS-Rollout significantly outperforms the baselines in tasks where long-horizon planning matters} (RLHF and Chess Endgame). When searching on shallow trees, they are robust enough to maintain comparable accuracy to the baselines.

\begin{figure*}[t]
    \centering
    \includegraphics[width=0.9\linewidth]{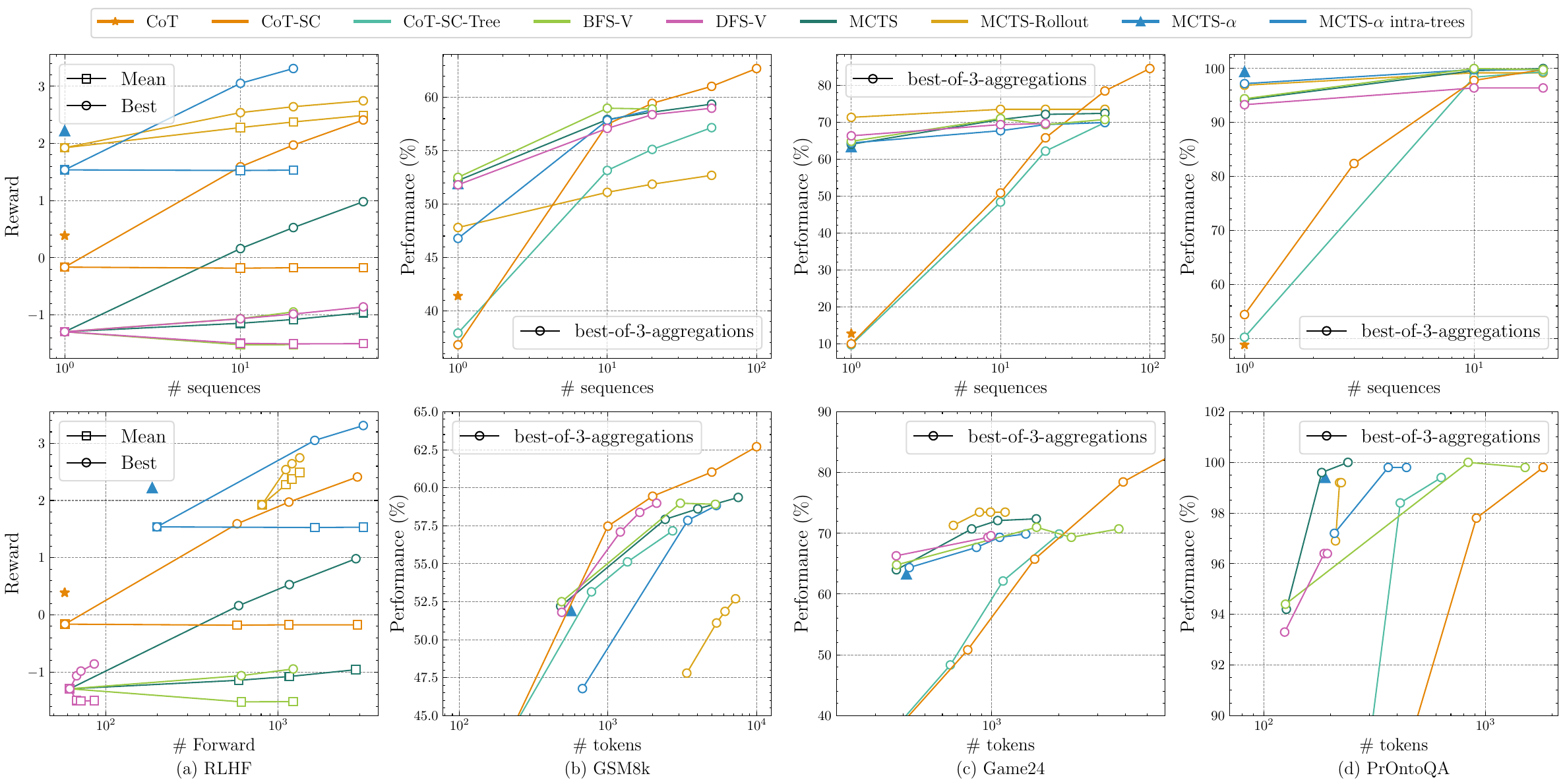}
    \vspace{-2mm}
    \caption{Aggregation results 
 for four tasks w.r.t. number of sequences\revision{(Path@$N$)} on the 1st row and the number of tokens on the 2nd row. TS-LLM benefits from aggregation but struggles to scale in small-scale problems.}
    \label{fig:2x4_res}
    \vspace{-5mm}
\end{figure*}

Despite the superiority of TS-LLM, we argue that the Path@1/Path@$N$ metric may not be reasonable. We also include the number of computations used in Path@1 generation (average number of tokens in sentence-level and average number of forward computation in token-level of solving one single problem). We refer readers to the \textbf{second} row of Fig~\ref{fig:2x4_res} for Path@$N$ result, with token/forward number as the x-axis. TS-LLM variants consume much more computation than CoT, making the comparison unfair.

To enable a fair comparison, in the bottom part of Table~\ref{tab:path1_result}, we show the "Equal-Token" results that try to compare results by controlling a similar scale of computation consumption with Path@1 TS-LLM. First, we provide additional baselines, CoT-SC with two aggregation methods: majority-vote (MAJ) and ORM-vote (denoted as ORM, and it utilizes the learned ORM in \method). Under this situation, TS-LLM's advantages largely decrease when compared with CoT-SC$_\text{ORM}$, especially on GSM8K (only BFS greedy value search is the best). We are surprised to see that such simple algorithms can also have outstanding performance when compared fairly. Despite this, \textbf{most tree-search algorithms are still dominant in the rest four tasks given the larger search space (CoT-SC)}.
   
Besides, we also compare the behaviors of BFS-/DFS-V and MCTS when searching for multiple paths (aggregated by the ORM model) within a comparable range of computation consumptions. Comparing these 3 variants, MCTS is almost the best w.r.t. both performance and computation cost, this indicates \textbf{the importance of value back-propogation}. While comparing with the Path@1 results, MCTS-$\alpha$ and MCTS-Rollout achieve comparable accuracy in shallow-search problems (GSM8k, Game24, and ProntoQA), and dominate in deep-search ones (RLHF and Chess Endgame). It verifies the necessity of \textbf{Alphazero-style intermediate value back-propagation under deep-search problems}.

\subsubsection{Search Aggregation (Sec \ref{sec-3.2.3-multi_search_aggregation})}
In Fig.~\ref{fig:2x4_res}, we demonstrate the mean/max reward for the RLHF task and the best of 3 aggregation results for GSM8K, Game24 and ProntoQA. We measure the performance of aggregation w.r.t path number and token consumption. 

From the figure, we mainly summarize two conclusions: First, \textbf{Most TS-LLM variants benefit from aggregation and can show large strengths compared with other baselines.} CoT-SC only beats TS-LLM in GSM8k with the same token size, mainly because of its larger search space. We refer the readers to Appendix \ref{apx: node_expansion} for additional results on GSM8K and ProntoQA. Second, \textbf{tree-search algorithms' aggregation benefits less than CoT-SC in small-scale problems.} In GSM8K and Game24, TS-LLM struggles to improve under large aggregation numbers. We believe this is because of: (1) The search space gap between CoT-SC and tree-search algorithms. Tree-search algorithms inherently explore fewer sentences, which is validated by comparing token consumption between CoT-SC-Tree@50 and CoT-SC@50. 
(2) Tree-search algorithms already leverage the value function and ORM, the benefits of aggregation with the ORM again become less obvious. 

In all, the scalability of tree-search aggregation is an open question that is worth further exploration in future work.

\subsubsection{TS-LLM for training LLM (Sec.~\ref{sec-3.3-trian_with_TS})}

  
\begin{table}[t]
\small
    \caption{Iterative update results. $\theta_{0},\phi_{0}$ are the old parameters while $\theta_{1},\phi_{1}$ are the new ones. TS-LLM can boost performance by training LLM policy, value, or both.}
    \label{tab:iterative_update}
    \vspace{-3mm}
    \small
        \centering
        \begin{tabular}{c c c c c}
            \toprule
            Task & Method & Policy & Value & Accuracy(\%) \\
            \midrule 
            \multirow{8}{*}{GSM8K} & \multirow{4}{*}{Greedy} & $\pi_{\theta_0}$ & - & 41.4 \\
            &  & $\pi_{\theta_1}$ & - & \textbf{47.9} \\
            &  & RFT-50 & - & 47.0 \\
            &  & RFT-100 & - & 47.5 \\
            \cmidrule{2-5} 
            & \multirow{4}{*}{MCTS-$\alpha$}& $\pi_{\theta_0}$ & $\{v, \hat{r}\}_{\phi_0}$ & 51.9  \\
            & & $\pi_{\theta_0}$ & $\{v, \hat{r}\}_{\phi_1}$ & 53.2\\
            & & $\pi_{\theta_1}$ & $\{v, \hat{r}\}_{\phi_0}$ & 54.1 \\
            & & $\pi_{\theta_1}$ & $\{v, \hat{r}\}_{\phi_1}$ & \textbf{56.5 } \\
            \midrule
            \multirow{8}{*}{RLHF}&         \multirow{4}{*}{Greedy} & $\pi_{\theta_0}$ & - & 0.39  \\
            & & $\pi_{\theta_1}$ & - & 1.87  \\
            & & RFT N=5 & - & 1.16  \\
            & & PPO & - & \textbf{2.53 } \\
            \cmidrule{2-5} 
            &\multirow{4}{*}{MCTS-$\alpha$}& $\pi_{\theta_0}$ & $\{v, \hat{r}\}_{\phi_0}$ & 2.22  \\
            & & $\pi_{\theta_0}$ & $\{v, \hat{r}\}_{\phi_1}$ & 2.48  \\
            & & $\pi_{\theta_1}$ & $\{v, \hat{r}\}_{\phi_0}$ & 2.53  \\
            & & $\pi_{\theta_1}$ & $\{v, \hat{r}\}_{\phi_1}$ & \textbf{2.67} \\
        \bottomrule
        \end{tabular}
    \hfill
    \vspace{-7mm}
\end{table}

We conduct initial experiments for one iterative update in GSM8k and RLHF alignment task. We utilize MCTS-$\alpha$ with old policy $\pi_{\theta_0}$, value $v_{\phi_0}$ and ORM $\hat{r}_{\phi_0}$, to sample answers on the training dataset as an augmentation to the origin one. 
It will be further used to finetune these models to  ($\pi_{\theta_1}, v_{\phi_1}, \hat{r}_{\phi_1}$). We include two baselines, RFT~\citep{yuan2023scaling}, which utilizes rejection sampling to finetune the policy, with different sampling numbers $k$ or top $N$, and PPO~\citep{schulman2017proximal} for the RLHF task. Note that PPO conducts multiple iterative updates. It is not fully comparable to our method and we only add it for completeness. Refer to Appendix~\ref{appendix:iterative_update} and \ref{appendix:iterative_update-train} for more experimental details and Appendix~\ref{app:value_ablation-iterative} for an ablation about how to use the collected data to optimize the the value and ORM.

In Table.~\ref{tab:iterative_update}, we list results of iterative update on the GSM8K and RLHF, covering greedy decoding and MCTS-$\alpha$ over all policy and value combinations. 
\textbf{Our empirical results validate that TS-LLM can further train LLM policy, value and ORM, boosting performance with the new policy $\pi_{\theta_1}$, new value and ORM $\{v, \hat{r}\}_{\phi_1}$, or both $(\pi_{\theta_1}, \{v, \hat{r}\}_{\phi_{1}})$} in CoT greedy decoding and MCTS-$\alpha$. 
$\pi_{\theta_1}$'s greedy performance is even slightly better than RFT which is specifically designed for GSM8k. We believe by further extending TS-LLM to multi-update, we can make it more competitive though currently $\pi_{\theta_{1}}$ still cannot beat PPO-based policy.

\subsubsection{Ablation studies}

\textbf{How to learn value function} We investigate data collection and training paradigms for value function and ORM in TS-LLM. In Table~\ref{tab:llm_train_value}, we investigate the influence of data amount and diversity by training with mixed data uniformly sampled from checkpoints of all SFT epochs (\textit{mixed}); data purely sampled from the last checkpoint (\textit{pure}); 1/3 data of the \textit{pure} setting (\textit{pure, less}). The results of CoT-SC$_\text{ORM-vote}$@10 underscore the diversity of sampled data in learning a better ORM. The Path@1 results of 3 TS-LLM variants show that the amount of sampled data is of great importance.  
Our final conclusion is that 
\textbf{collecting a diverse dataset is better for the ORM and collecting as much data as possible is better for value function training.}
And we also leave an ablation about how to use the collected data to optimize the value and ORM for iterative update in Appendix~\ref{app:value_ablation-iterative}.

\begin{table}[t]
\small
    \centering
    \caption{Path@1 results with different training settings.}
    \label{tab:llm_train_value}
    \begin{tabular}{ccc}
        \toprule
        Search Algorithms & Training Setting & Accuracy(\%) \\
        \midrule
        \multirow{3}{*}{CoT-SC@10@{ORM}} & Pure, Less  & 55.5 $\pm$ 0.6 \\
          & Pure & 55.3 $\pm$ 0.5\\
         & Mixed &  \textbf{55.9 $\pm$ 0.7} \\
        \midrule
         \multirow{3}{*}{BFS} & Pure, Less   & 50.0 $\pm$ 0.3 \\
         & Pure & \textbf{52.7 $\pm$ 0.8}\\
         & Mixed & 52.5 $\pm$ 1.3\\
         \midrule
         \multirow{3}{*}{MCTS-$\alpha$} & Pure, Less   & 49.7 $\pm$ 1.1 \\
         & Pure & \textbf{52.7 $\pm$ 0.8}\\
         & Mixed & \textbf{51.9 $\pm$ 0.6}\\
        \bottomrule
    \end{tabular}
\end{table}

\textbf{Search space and search width}
As discussed in Sec.~\ref{sec:method-problem_formulation}, the search space is limited \revision{by maximum tree width}. We demonstrate the influence introduced by different tree constructions on Game24 with different node expansion sizes. 
Specifically, we set the number of maximal expanded node sizes as 6, 20, and 50. Table~\ref{tab:ablation_tree_size} lists the \revision{Path@1} performance and the number of tokens generated comparing TS-LLM's variants, CoT-SC and CoT-SC-Tree. The almost doubled performance boost from 43.8 to 80.7 indicates the impact of different expansion sizes and tree-width, improving TS-LLM's performance upper bound. Refer to Appendix~\ref{apx: node_expansion} for additional results on GSM8K and ProntoQA.
\begin{table}[t]
    \centering 
    \small
    \caption{Game24 Path@1 result with different tree-width. Larger search space leads to much better performance.}\label{tab:ablation_tree_size}
    \vspace{-0.7mm}
     \begin{tabular}{c c c c}
     \toprule
      & \multicolumn{3}{c}{Performance(\%) / \# tokens} \\
     \cmidrule{2-4}
     Method & {width=6} & {width=20} & {width=50} \\
     \midrule
     MCTS-$\alpha$  & 41.6  & 63.3 &  74.5  \\ 
     MCTS-Rollout        & 43.8  & 71.3 &  80.7  \\
     BFS-V            & 43.2  & 64.8  & 74.6  \\
     CoT-SC-Tree@10 & 38.8  & 48.3  & 48.3  \\
     CoT-SC@10 & -  & -  & 52.9  \\
     \bottomrule
    \end{tabular}
    \vspace{-3mm}
\end{table}

%% file: sections/5-conclusion.tex
\vspace{-5pt}
\section{Conclusion}
In this work, we propose \method, an LLM inference and training framework guided by Alphazero-like tree search that is generally versatile for different tasks and scaled to token-level expanded tree spaces. Empirical results validate that \method~can enhance LLMs decoding and serve as a new training paradigm. We leave further discussion of our limitation and future work in Appendix~\ref{appendix:limitation_and_future}.

%% file: sections/appendix.tex
\appendix
\addcontentsline{toc}{section}{Appendix}
\section{More Related Work and Comparisons}
\label{app:comparsion-related_work}
Here we discuss the differences between \method~and relevant work mentioned in 
Sec~\ref{sec:related_work} in detail.

Recent efforts have been made to investigate non-linear reasoning structures such as trees\citep{jung2022maieutic, zhu2022mctscooperative, xu2023notrain, xie2023decomposition, yao2023tot, hao2023rap}.
Approaches for searching on trees with LLM's self-evaluation have been applied to find better reasoning paths, e.g. beam search in \citet{xie2023decomposition}, depth-/breadth-first search in \citet{yao2023tot} and Monte-Carlo Tree Search in \citep{hao2023rap}.
Compared with these methods, \method~is a tree search guided LLM decoding framework with a learned value function, which is more generally applicable to reasoning tasks and other scenarios like RLHF alignment. 
\method~includes comparisons between different tree search approaches, analysis of computation cost, and shows the possibility of improving both the language model and value function iteratively. 
The most relevant work, CoRe \citep{zhu2022mctscooperative}, proposes to finetune both the reasoning step generator and learned verifier for solving math word problems using MCTS for reasoning decoding which is the most relevant work to ours. 
Compared with CoRe, in this work \method~distinguishes itself by: 
\\
\textbf{1.} \method~is generally applicable to a wide range of reasoning tasks and text generation tasks under general reward settings, from sentence-level trees to token-level trees. But CoRe is proposed for Math Word Problems and only assumes a binary verifier (reward model).
\\
\textbf{2.} In this work, we conduct comprehensive comparisons between popular tree search approaches on reasoning, planning, and RLHF alignment tasks. We fairly compare linear decoding approaches like CoT and CoT-SC with tree search approaches w.r.t. computation efficiency.
\\
\textbf{3.} \method~demonstrates potentials to improve LLMs' performance of direct decoding as well as tree search guided decoding, while in CoRe the latter cannot be improved when combining the updated generator (language model policy) with the updated verifier (value function) together.

Other related topic:\\
\textbf{Search guided decoding in LLMs}
Heuristic search and planning like beam search or MCTS have also been used in NLP tasks including machine translation \citep{leblond2021machine} and code generation \citep{zhang2023planning, matulewicz2022inductive}. 
During our preparation for the appendix, we find a concurrent work \citep{liu2023making} which is proposed to guide LLM's decoding by reusing the critic model during PPO optimization to improve language models in alignment tasks. Compared with this work, \method~focuses on optimizing the policy and value model through tree search guided inference and demonstrates the potential of continuously improving the policy and value models. And \method~is generally applicable to both alignment tasks and reasoning tasks by conducting search on token-level actions and sentence-level actions.
\section{Limitation and future work}
\label{appendix:limitation_and_future}
Currently, our method \method~still cannot scale to really large-scale scenarios due to the extra computation burdens introduced by node expansion and value evaluation. Additional engineering work such as key value caching is required to accelerate the tree-search. In addition, we do not cover all feasible action-space designs for tree search and it is flexible to propose advanced algorithms to automatically construct a tree mixed with both sentence-level expansion and token-level expansion, etc. We leave such exploration for future work. For MCTS aggregation, the current method still struggles to improve under large aggregation numbers. some new algorithms that can encourage multi-search diversity might be needed.  Currently, we are still actively working on scaling up our method both during inference and training (especially multi-iteration training). 

\section{Backgrounds and details of each tree-search algorithms in \method}
\subsection{Preliminaries of Monte Carlo Tree-search Algorihtms}
\label{apx:ts_algos}
Once we build the tree, we can use various search algorithms to find a high-reward trace. However, it's not easy to balance between exploration and exploitation during the search process, especially when the tree is sufficiently deep. 
Therefore we adopt Monte Carlo Tree Search(MCTS) variants as choices for strategic and principled search. 
Instead of the four operations in traditional MCTS \citep{kocsis2006mcts-bandit, coulom2006mcts-efficient}, we refer to the search process in AlphaZero \citep{silver2017alphazero} and introduce 3 basic operations of a standard search simulation in it as follows, when searching actions from current state node $s_0$:

\textbf{\textit{Select}} It begins at the root node of the search tree, of the current state, $s_0$, and finishes when reaching a leaf node $s_L$ at timestep $L$. At each of these $L$ timesteps(internal nodes), an action(child node) is selected according to $a_t = \displaystyle \arg\max\limits_a \left( Q(s_t, a) + U(s_t, a) \right)$ where $U(s_t, a)$ is calculated by a variant of PUCT algorithm \citep{rosin2011puct}:
\begin{equation}
    U(s, a) = c_\text{puct}\cdot \pi_\theta(s, a) \frac{\sqrt{\sum_b N(s, b)}}{1 + N(s, a)}
\end{equation}
where $N(s, a)$ is the visit count of selecting action $a$ at node $s$, and $c_\text{puct} = \log((\sum_b N(s, b) + c_\text{base} + 1) / c_\text{base}) + c_\text{init}$ is controlled by visit count and two constants.
This search control strategy initially prefers actions with high prior probability and low visit count, but asymptotically prefers actions with high action-value.

\textbf{\textit{Expand and evaluate}} After encountering a leaf node $s_L$ by \textit{select}, if $s_L$ is not a terminal node, it will be expanded by the language model policy. The state of the leaf node is evaluated by the value network, noted as $v(s_L)$. If $s_L$ is a terminal node, if there is an oracle reward function $R$, then $v(s_L) = R(s_L)$, otherwise, in this paper, we use an ORM $\hat{r}$ as an approximation of it.

\textbf{\textit{Backup}} After \textit{expand and evaluate} on a leaf node, backward the statistics through the path $s_L, s_{L-1}, \ldots, s_0$, for each node, increase the visit count by $N(s_t, a_t) = N(s_t, a_t)+1$, and the total action-value are updated as $W(s_t, a_t) = W(s_t, a_t) + v(s_L)$, the mean action-value are updated as $Q(s_t, a_t) = W(s_t, a_t) / N(s_t, a_t)$.

\subsection{Comparison of tree-search algorithms in \method}

In this paper, we introduce three variants of MCTS based on the above basic operations. \revision{Among the 3 variants, MCTS-$\alpha$ is closer to AlphaZero\citep{silver2017alphazero} and MCTS is closer to traditional Monte-Carlo Tree Search\citep{kocsis2006mcts-bandit}. While MCTS-Rollout is closer to best-first search or A*-like tree search.}

\revision{The major difference between the first three MCTS variants and BFS-V/DFS-V adopted from the ToT paper\citep{yao2023tot} is that the first three MCTS variants will propagate (i.e. the \textit{backup} operation) the value and visit history information through the search process. 
MCTS-$\alpha$ and MCTS-Rollout bacpropagate after the \textit{expand} operation or visiting a terminal node. MCTS backpropagates the information only after visiting a terminal node.
For DFS-V, the children of a non-leaf node is traversed in a non-decreasing order by value. For efficient exploration, we tried 2 heuristics to prune the subtrees, (1) drop the children nodes with low value by \textit{prune\_ratio}. (2) drop the children nodes lower than a \textit{prune\_value}.} 

\section{Extra Experiments and Discussions}
\subsection{Different Value Training of iterative update}
\label{app:value_ablation-iterative}

\begin{table}[h]
    \centering
    \small
    \caption{Different value training for iterative update on GSM8k}
    \label{tab:value_ablation-iterative}
    \begin{tabular}{c c c c}
        \toprule
        Method & Policy & Value & Performance(\%) \\
        \midrule 
        \multirow{3}{*}{MCTS-$\alpha$} & $\pi_{\theta_0}$ & $\{v, \hat{r}\}_{\phi_0}$ & 51.9 $\pm$ 0.6 \\

         & $\pi_{\theta_0}$ & $\{v, \hat{r}\}_{\phi_1}^\text{RL}$ & 52.0 $\pm$ 0.5 \\
        
         & $\pi_{\theta_0}$ & $\{v, \hat{r}\}_{\phi_1}$ & 53.2 $\pm$ 0.3 \\
        \midrule
        MCTS-$\alpha$ & $\pi_{\theta_1}$ & $\{v, \hat{r}\}_{\phi_0}$ & 54.1 $\pm$ 0.9 \\

        MCTS-$\alpha$ & $\pi_{\theta_1}$ & $\{v, \hat{r}\}_{\phi_1}^\text{RL}$ & 55.2 $\pm$ 1.2 \\
        
        MCTS-$\alpha$ & $\pi_{\theta_1}$ & $\{v, \hat{r}\}_{\phi_1}$ & \textbf{56.5 $\pm$ 0.6} \\
        \bottomrule
    \end{tabular}
\end{table}
To figure out the best way of training the value function during iterative update. We also compare MCTS-$\alpha$ in Table~\ref{tab:value_ablation-iterative}. 
We train value and ORM in two paradigms, one ($\{v, \hat{r}\}_{\phi_1}$) is optimized from the initial weights and mixture of old and new tree-search data; another($\{v, \hat{r}\}_{\phi_1}^\text{RL}$) is optimized from $\{v, \hat{r}\}_{\phi_0}$ with only new tree-search data. This is called RL because training critic model in RL utilizes a similar process of continual training.
The results show that $\{v, \hat{r}\}_{\phi_1}$ outperforms $\{v, \hat{r}\}_{\phi_1}^\text{RL}$ on both old and new policy when conducting tree search, contrary to the normal situation in traditional iterative RL training.


  

\subsection{Results of different node expansion on tasks}
\label{apx: node_expansion}
\revision{Table~\ref{tab:ablation_tree_size-gsm8k}, Table~\ref{tab:ablation_tree_size-game24} and Table~\ref{tab:ablation_tree_size-prontoqa} show the path@1 results of MCTS-$\alpha$, MCTS-Rollout, BFS-V under different numbers of \textit{tree-max-width} $w$ on GSM8k, Game24 and ProntoQA. And we also show the results of CoT-SC-Tree$_\text{ORM}$@10 and CoT-SC$_\text{ORM}$@10, which are aggregated by ORM-vote. The results are conducted under 3 seeds and we show the average value and standard deviation.}

Let us first clarify how we choose the specific tree max width. For \textit{tree-max-width} $w$ in GSM8k, Game24, and ProntoQA, we first start with an initial value $w=6$. Then by increasing it (10 in GSM8K, 20 in Game 24, and 10 in ProntoQA), we can see the trends in performance and computation consumption. In ProntoQA/GSM8K, the performance gain is quite limited while the performance gain is quite large in Game24. So at last, we in turn try smaller \textit{tree-max-width} in GSM8K/ProntoQA (3) and also try even larger \textit{tree-max-width} (50) in Game24. Our final choice of $w$ (in Table \ref{tab:exp-task_setup}) is based on the trade-off between the performance and the computation consumption. Currently, our selection is mainly based on empirical trials and it might be inefficient to determine the appropriate \textit{tree-max-width} $w$. We think this procedure can be more efficient and automatic by comparing it with the results of CoT-SC on multiple samples to balance the tradeoff between performance and computation consumption. Because CoT-SC examples can already provide us with information about the model generation variation and diversity. We can also leverage task-specific features, e.g. in Game24, the correctness of early steps is very important, so a large $w$ can help to select more correct paths from the first layers on the search trees.

For the analysis of the results in Table~\ref{tab:ablation_tree_size-gsm8k}, Table~\ref{tab:ablation_tree_size-game24} and Table~\ref{tab:ablation_tree_size-prontoqa}. We can mainly draw two conclusions aligned with that in Q1/Q2 in the main paper. 

\revision{First, the overall trend that larger search space represents better tree-search performance still holds. For most tree-search settings, larger \textit{tree-max-width} $w$ and search space bring in performance gain. The only exception happens at MCTS-Rollout on GSM8k decreases when the \textit{tree-max-width} $w=10$, this is due to the limitation of computation(limitation on the number of generated tokens which is $51200$ per problem) is not enough in wider trees which results in more null answers. Despite the gain, the number of generated tokens also increases as the tree-max-width $w$ becomes larger.}

\revision{Secondly, the conclusions in the main paper about comparing different search algorithms still hold. BFS performs pretty well in shallow search problems like (GSM8K/Game24). Though we can still see MCTS-$\alpha$ and MCTS-Rollout improve by searching in large \textit{tree-max-width} (such as Game24 expanded by 50), the performance gain is mainly attributed to the extra token consumption and is quite limited. For deeper search problems like ProntoQA (15) and RLHF (64), the performance gap is obvious and more clear among all expansion widths. This aligns with our conclusion in Q1's analysis.}

\begin{table}[h!]
    \centering 
    \caption{\revision{Path@1 metric on GSM8k with different node size.}}
    \label{tab:ablation_tree_size-gsm8k}
    \vspace{-1.7mm}
     \begin{tabular}{c cc cc cc}
     \toprule
     \multirow{2}{3em}{Method} & \multicolumn{6}{c}{Performance(\%) / \# tokens} \\
     \cmidrule{2-7}
     & \multicolumn{2}{c}{expand by 3} & \multicolumn{2}{c}{expand by 6} & \multicolumn{2}{c}{expand by 10} \\
     \midrule
     MCTS-$\alpha$  & 49.2 $\pm$ 0.04 & 460 & 51.9 $\pm$ 0.6 & 561 & 51.7 $\pm$ 0.5 & 824 \\ 
     MCTS-Rollout   & 47.2 $\pm$ 0.8 & 856 & 47.8 $\pm$ 0.8 & 3.4k & 45.9 $\pm$ 0.9 & 7.1k \\
     BFS-V          & 49.1 $\pm$ 0.8 & 260 & 52.5 $\pm$ 1.3 & 485 & 52.2 $\pm$ 0.9 & 778 \\
     CoT-SC-Tree$_\text{ORM}$@10 & 52.4 $\pm$ 1.2 & 604 & 54.6 $\pm$ 0.7 & 780 & 54.5 $\pm$ 1.1 & 857 \\
     CoT-SC$_\text{ORM}$@10 & - & - & - & - & 56.4 $\pm$ 0.6 & 1.0k \\
     \bottomrule
    \end{tabular}
\end{table}

\begin{table}[h!]
    \centering 
    \caption{\revision{Path@1 metric on Game24 with different node size.}}
    \label{tab:ablation_tree_size-game24}
    \vspace{-1.7mm}
     \begin{tabular}{c cc cc cc}
     \toprule
     \multirow{2}{3em}{Method} & \multicolumn{6}{c}{Performance(\%) / \# tokens} \\
     \cmidrule{2-7}
     & \multicolumn{2}{c}{expand by 6} & \multicolumn{2}{c}{expand by 20} & \multicolumn{2}{c}{expand by 50} \\
     \midrule
     MCTS-$\alpha$  & 41.6 $\pm$ 0.8 & 243 & 63.3 $\pm$ 1.9 & 412 & 74.5 $\pm$ 0.7 & 573 \\ 
     MCTS-Rollout   & 43.8 $\pm$ 5.3 & 401 & 71.3 $\pm$ 2.5 & 670 & 80.7 $\pm$ 1.5 & 833 \\
     BFS-V          & 43.2 $\pm$ 2.0 & 206 & 64.8 $\pm$ 2.9 & 370 & 74.6 $\pm$ 0.5 & 528 \\
     CoT-SC-Tree$_\text{ORM}$@10 & 38.8 $\pm$ 2.0 & 508 & 48.3 $\pm$ 3.0 & 656 & 48.3 $\pm$ 4.2 & 707 \\
     CoT-SC$_\text{ORM}$@10 & - & - & - & - & 52.9 $\pm$ 2.1 & 0.8k \\
     \bottomrule
    \end{tabular}
\end{table}

\begin{table}[h!]
    \centering 
    \caption{\revision{Path@1 metric on ProntoQA with different node size.}}
    \label{tab:ablation_tree_size-prontoqa}
    \vspace{-1.7mm}
     \begin{tabular}{c cc cc cc}
     \toprule
     \multirow{2}{3em}{Method} & \multicolumn{6}{c}{Performance(\%) / \# tokens} \\
     \cmidrule{2-7}
     & \multicolumn{2}{c}{expand by 3} & \multicolumn{2}{c}{expand by 6} & \multicolumn{2}{c}{expand by 10} \\
     \midrule
     MCTS-$\alpha$  & 94.1 $\pm$ 0.1 & 151 & 99.4 $\pm$ 0.2 & 190 & 99.8 $\pm$ 0.2 & 225 \\ 
     MCTS-Rollout   & 85.9 $\pm$ 0.8 & 151 & 96.9 $\pm$ 0.6 & 210 & 99.3 $\pm$ 0.4 & 264 \\
     BFS-V          & 83.7 $\pm$ 1.0 & 105 & 94.4 $\pm$ 0.3 & 126 & 97.6 $\pm$ 0.3 & 145 \\
     CoT-SC-Tree$_\text{ORM}$@10 & 91.9 $\pm$ 0.8 & 290 & 98.2 $\pm$ 0.4 & 417 & 99.1 $\pm$ 0.1 & 494 \\
     CoT-SC$_\text{ORM}$@10 & - & - & - & - & 98.0 $\pm$ 0.7 & 0.9k \\
     \bottomrule
    \end{tabular}
\end{table}

\subsection{Wall-time and Engineering challenges}
\label{appendix:wall-time-and-engineering}

\revision{
Table~\ref{tab:wall-time-gsm8k}, Table~\ref{tab:wall-time-game24}  and Table~\ref{tab:wall-time-prontoqa} show the wall-time of running different tree search algorithms implemented in \method~, searching for one answer per problem (i.e. path@1). 
We also show the wall-time of CoT greedy decoding and CoT-SC@10 with ORM aggregation as comparisons. 
We record the wall-time of inferencing over the total test dataset of each task.}

\revision{The experiments were conducted on the same machine with 8 NVIDIA A800 GPUs, the CPU information is Intel(R) Xeon(R) Platinum 8336C CPU @ 2.30GHz.}

\revision{
We can find that the comparisons of wall-time within all the implemented tree-search algorithms in \method~ are consistent with those of the number of generated tokens. 
However, compared to CoT greedy decoding, the wall-time results of most tree search algorithms in \method~are between two and three times of CoT greedy decoding's wall-time, excepting MCTS-Rollout runs for a very long time on GSM8k.
And when comparing the wall-time and number of generated tokens between the tree-search methods and CoT-SC$_\text{ORM}$@10, \method~is not as computationally efficient as CoT-SC decoding due to the complicated search procedures and extra computation introduced by calling value functions in the intermediate states.
}

\revision{There are two more things we want to clarify. Firstly, as we mentioned in Appendix \ref{appendix:limitation_and_future}, our current implementation only provides an algorithm prototype without specific engineering optimization. We find a lot of repeated computations are performed in our implementation when evaluating the child node's value. Overall, there still exists great potential to accelerate the tree-search process, which will be discussed in the next paragraph. We are continuously working on this (we will discuss the engineering challenges in the next paragraph). Secondly, this wall-time is just Path@1 result so they have different token consumptions, and DFS-V, BFS-V, and MCTS will degenerate into greedy value search as we mentioned before. We are also working on monitoring the time consumption for Path@N results so we can compare them when given the same scale of token consumption.}

\revision{\textbf{Engineering challenges and potentials.} Here we present several engineering challenges and potentials to increase the tree-search efficiency.}
\begin{itemize}[leftmargin=*]
    \item \textbf{Policy and Value LLM with the shared decoder.} Our current implementation utilizes separate policy and value decoders but a shared one might be a better choice for efficiency. If so, most extra computation brought by value evaluation can be reduced to simple MLP computation (from the additional value head) by reusing computation from LLM's policy rollout. It can largely increase the efficiency. We only need to care about the LLM's rollout computations under this setting.
    \item \textbf{KV cache and computation reuse.} KV cache is used in most LLM's inference processes such as the Huggingface transformer's \citep{wolf-etal-2020-transformers} generation function. It saves compute resources by caching and reusing previously calculated key-value pairs in self-attention. In the tree search problem, when expanding or evaluating a node, all preceding calculations for its ancestor nodes can be KV-cached and reused. However, because of the large state space of tree nodes, we cannot cache all node calculations since the GPU memory is limited and the communication between GPU/CPU is also inefficient (if we choose to store such cache in CPU). More engineering work is needed to handle the memory and time tradeoff.
    \item \textbf{Large-batch vectorization.} Currently, our node expansion and node evaluation are only vectorized and batched given one parent node. We may conduct batch inference over multiple parent nodes for large-batch vectorization when given enough computing resources.
    \item \textbf{Parallel tree-search over multi-GPUs.} Our implementation handles each tree over a single GPU. AlphaZero \citep{silver2017alphazero} leverages parallel search over the tree\citep{segal2010scalability}, using multi-thread search to increase efficiency. In LLM generation setting, the main bottleneck comes from the LLM inference time on GPU. Thus more engineering work is needed for conducting parallel tree-search over multi-GPUs.
    \item \textbf{Tree-Search with speculative decoding} Speculative decoding \citep{leviathan2023fast} is a pivotal technique to accelerate LLM inference by employing a smaller draft model to predict the target model's outputs. During the speculative decoding, the small LLM gives a generation proposal while the large LLM is used to evaluate and determine whether to accept or reject the proposal. This is similar to the tree-search process with value function pruning the sub-tree. There exists potential that by leveraging small LLMs as the rollout policy while large LLMs as the value function, we can also have efficient tree-search implementations.
\end{itemize}

\begin{table}[h!]
        \centering 
        \caption{Wall-time results on GSM8k}
        \label{tab:wall-time-gsm8k}
        \vspace{-1em}
        \begin{tabular}{c c c c}
        \toprule
        Method        & Overall Time(sec) & Average Time(sec) & \#Average Token\\
        \midrule
        CoT-Greedy    & 216.93   & 0.17  & 98   \\
        CoT-SC$_\text{ORM}$@10 & 479.03 & 0.37 & 1k   \\
        \midrule
        MCTS          & 378.13  & 0.29 & 486  \\
        MCTS-$\alpha$    & 527.31 & 0.41 & 561  \\
        MCTS-Rollout  & 2945.94 & 2.27 & 3.4k \\
        BFS-V           & 383.08 & 0.29 & 485  \\
        DFS-V           & 387.00 & 0.30 & 486  \\   
        \bottomrule
    \end{tabular}
\end{table}

\begin{table}[h!]
        \centering 
        \caption{Wall-time results on Game24}
        \label{tab:wall-time-game24}
        \vspace{-1em}
        \begin{tabular}{c c c c}
        \toprule
        Method        & Overall Time(sec) & Average Time(sec) & \#Average Token\\
        \midrule
        CoT-Greedy    & 44.88  & 0.15 &76   \\
        CoT-SC$_\text{ORM}$@10 & 86.53 & 0.29 & 0.8k \\
        \midrule
        MCTS          & 81.22 & 0.27 & 371  \\
        MCTS-$\alpha$    & 134.19& 0.45 & 412  \\
        MCTS-Rollout  & 193.18& 0.64 & 670  \\
        BFS-V           & 79.48& 0.26 & 369  \\
        DFS-V           & 80.46& 0.27 & 369 \\
        \bottomrule
    \end{tabular}
\end{table}

\begin{table}[h!]
        \centering 
        \caption{Wall-time results on ProntoQA}
        \label{tab:wall-time-prontoqa}
        \vspace{-1em}
        \begin{tabular}{c c c c}
        \toprule
        Method        & Overall Time(sec) & Average Time(sec) & \#Average Token\\
        \midrule
        CoT-Greedy    & 74.09   & 0.15 &77   \\
        CoT-SC$_\text{ORM}$@10 & 218.12 & 0.44 &0.8k \\
        \midrule
        MCTS          & 130.15 & 0.26 &125  \\
        MCTS-$\alpha$    & 236.26 & 0.47 & 190  \\
        MCTS-Rollout  & 238.62 & 0.48 &210  \\
        BFS-V           & 130.35 & 0.26 & 126  \\
        DFS-V           & 130.47 & 0.26 & 126  \\
        \bottomrule
    \end{tabular}
\end{table}

\subsection{Discussion about Shared LLM decoder for both policy and critic.}
\label{apx:shared_decoder}
As we mentioned in Appendix~\ref{appendix:wall-time-and-engineering},
using a shared decoder for the policy and value LLM might further improve the computation efficiency for the tree-search process. Therefore, we conducted an ablation to compare the model under the settings of a \textit{shared decoder} and the setting of \textit{separated decoder}s on Game24.

We first describe the training setting of both types of models we compared. For the setting of \textit{separated decoder}, we refer to Appendix~\ref{appendix:sft_value_training} for details about dataset and training hyperparameters.
For the setting of \textit{shared decoder}, we train the shared policy and value LLM with the same data used in the setting of \textit{separated decoder}. 
During training, a batch from the supervised finetuning (SFT) dataset and a batch from the value training dataset are sampled, the total loss of the shared policy and value LLM is computed by $\mathcal{L}_\text{total} = \mathcal{L}_\text{SFT} + 0.5 \cdot \mathcal{L}_\text{Value}$, where $\mathcal{L}_\text{SFT}$ is the cross entropy loss of predicting the next token in the groundtruth answer and $\mathcal{L}_\text{Value}$ is the Mean Square Error loss as we described in Equation~\ref{equation:value_loss}.
The training is conducted on 8 NVIDIA A800 GPUs, using a
cosine scheduler decaying from lr=2e-5 to 0.0 with a warmup ratio of 0.03, a batch size of 128 for the supervised finetuning dataset and a batch size of 128 for the value training dataset. And we trained the shared decoder model for 3 epochs. This training setting is the same as used in the \textit{separated decoder} setting.

We will compare these two types of model in the perspective of performance and computation efficiency.
All tree-search algorithms are conducted under the same hyperparameters as those in Table~\ref{tab:path1_result} on Game24 in which the tree-max-width $w$ is set to 20.

Table~\ref{tab:ablation_shared-decoder} shows the comparisons of the two types of model on performance. Though the performance of CoT (CoT greedy decoding) of \textit{shared decoder} model increases from 12.7 to 16.3, the number of tokens generated per problem also increases greatly from 76 to 166. By checking the models' outputs, we find the \textit{shared decoder} model doesn't always obey the rules of Game24(There are only 4 steps of calculations and each number must be used exactly once). It usually outputs multiple steps, more than the four steps required in Game24. This might be regarded as hallucination problem which happens more frequently than in the \textit{separated decoder} model. 
For the results of CoT-SC-Tree$_\text{ORM}$@10 (search by LLM's prior on trees and aggregated by ORM-vote), we observe close results of CoT-SC$_\text{ORM}$@10
Meanwhile, for the results of MCTS-$\alpha$, MCTS-Rollut, BFS-V and CoT-SC-Tree$_\text{ORM}$@10, there is only a small difference between the performance of the two models. We also observe an increase in the number of token generated per problem. This is because the \textit{shared decoder} model is prone to output more invalid answers than the \textit{separated decoder} model. Therefore, there are more distinct actions proposed in the last layers of the trees.

Next, we show some preliminary comparisons in Table~\ref{tab:ablation_shared-decoder-time} on the computation efficiency of the \textit{separated decoder} model and the \textit{shared decoder} model, from the results of expanding a tree node and evaluating its children. Specifically, Table~\ref{tab:ablation_shared-decoder-time} presents the node expansion time and value calculation time with/without KV Cache under token-level and sentence-level situations. For the token-level node, we set $w=50$ while for the sentence-level node, we set $w=20$. The results successfully present that a shared decoder can largely increase the computational efficiency for value estimation (20x in the token-level setting and 9x in the sentence-level setting).

In all, in this section, we initially conduct explorations on leveraging shared policy and value LLM decoder. The result proves the potential of computational efficiency for the shared structure. However, more work is needed to help the stability of policy/value performance.

\begin{table}[h!]
    \centering 
    \caption{\revision{Comparisions of separated/shared LLM decoder policy and critic models on Game24}}
    \label{tab:ablation_shared-decoder}
    \vspace{-1.7mm}
     \begin{tabular}{c cc | cc}
     \toprule
     \multirow{2}{3em}{Method} & \multicolumn{4}{c}{Performance(\%) / \# tokens} \\
     \cmidrule{2-5}
     & \multicolumn{2}{c}{Separated Decoder} & \multicolumn{2}{c}{Shared Decoder}\\
     \midrule
     CoT            & 12.7 & 76 & 16.3 & 166 \\
     CoT-SC$_\text{ORM}$@10 & 52.9 $\pm$ 2.1 & 0.8k & 52.8 $\pm$ 2.4 & 1.9k  \\
     \midrule
     MCTS-$\alpha$  & 63.3 $\pm$ 1.9 & 412 & 64.1 $\pm$ 1.3 & 561  \\ 
     MCTS-Rollout   & 71.3 $\pm$ 2.5 & 670 & 70.6 $\pm$ 0.4 & 855  \\
     BFS-V          & 64.8 $\pm$ 2.9 & 370 & 63.0 $\pm$ 1.0 & 495  \\
     CoT-SC-Tree$_\text{ORM}$@10 & 48.3 $\pm$ 3.0 & 656 & 45.5 $\pm$ 2.0 & 745  \\
     \bottomrule
    \end{tabular}
\end{table}

\begin{table}[h!]
    \centering 
    \caption{\revision{Time (seconds) on policy expansion and value evaluation for a single tree node. When using a shared LLM decoder for policy and value LLM, we can use KV Cache for value calculation. It is much more efficient than a separate value decoder without KV cache.}}
    \label{tab:ablation_shared-decoder-time}
    \vspace{-1.7mm}
     \begin{tabular}{c ccc}
     \toprule
    Node Type & Policy Expansion & Value with Cache & Value without Cache \\
    \midrule 
    Token-Level & 0.067 & 0.074 & 2.02 \\
     \midrule 
    Sentence-Level & 0.165 & 0.122 & 1.03 \\
     \bottomrule
    \end{tabular}
\end{table}

\section{Experiment Details}
\label{appendix:experiment_details}
\subsection{Task setups}
\label{appendix:task_setup}
\textbf{GSM8k} 
GSM8k \citep{cobbe2021gsm8k} is a commonly used numerical reasoning dataset, Given a context description and a question, it takes steps of mathematical reasoning and computation to arrive at a final answer. There are about 7.5k problems in the training dataset and 1.3k problems in the test dataset.
\\
\textbf{Game24}
We also test our methods on Game24\citep{yao2023tot} which has been proven to be hard even for state-of-the-art LLMs like GPT-4. Each problem in Game24 consists of 4 integers between 1 and 13. And LLMs are required to use each number exactly once by $(+-\times\div)$ to get a result equal to $24$
We follow \citet{yao2023tot} by using a set of 1362 problems sorted from easy to hard according to human solving time. We split the first 1k problems as the training dataset and the last 362 hard problems as the test dataset.
For each problem in the training dataset, we collect data for SFT by enumerating all possible correct answers.
\\
\textbf{PrOntoQA}
PrOntoQA \citep{saparov2022prontoqa} is a typical logical reasoning task in which a language model is required to verify whether a hypothesis is true or false given a set of facts and logical rules. There are 4k problems in the training dataset and 500 problems in the test dataset.
\\
\textbf{RLHF} We choose a synthetic RLHF dataset \cite{rlhf_data}\footnote{https://huggingface.co/datasets/Dahoas/synthetic-instruct-gptj-pairwise} serving as the query data. We split the dataset to 30000/3000 as training and test set respectively. For the reward model, we choose reward-model-deberta-v3-large-v2\footnote{https://huggingface.co/OpenAssistant/reward-model-deberta-v3-large-v2} from OpenAssistant, which is trained from several RLHF datasets.
\\
\textbf{Chess Endgame} Chess Endgame is introduced by \citet{abdulhai2023lmrl}. 
Chess endgames provide a simpler and more goaldirected variation of the chess task. 
A classic theoretical endgame position consists of a position where the only pieces on the board are the two kings and the queen. 
Although the board position appears simple, a sequence of carefully calculated moves is required to win. We followed the setting of \citet{abdulhai2023lmrl} use an opponent of StockFish whose Elo is 1200. We modified the environment dynamics so that the agent fails as long as he makes an illegal move.

\subsection{SFT and value training details}
\label{appendix:sft_value_training}

\textbf{SFT in GSM8k, Game24 and PrOntoQA}: 
For GSM8k, Game24 and PrOntoQA, we first train LLaMA2-7b on the training dataset The training is conducted on 8 NVIDIA A800 GPUs, using a cosine scheduler decaying from lr=2e-5 to 0.0 with a warmup ratio of 0.03, batch size 128 for 3 epochs. For GSM8k and Game24 we use the checkpoint at the last epoch as the direct policy in experiments, while for PrOntoQA we use the checkpoint at the 1st epoch since the others overfit.

\textbf{Value training in GSM8k, Game24 and PrOntoQA}:
Then we train the value function on the data rollout by the SFT policy. In GSM8k and Game24, For each model checkpoints of 3 epochs during SFT, we first collect 100 outputs per problem in the training dataset, then duplicate the overlapped answers, labeled each answer with our training set outcome reward ocracle. For data sampled by ech model checkpoint, we subsample 17 answers per problem, which is in total at most 51 answers per problem after deduplication.
In PrOntoQA, we only sample 50 answers per problem with the first epoch model checkpoint and then do deduplication.

The value functions are trained in the same setting as supervised finetuning. 
We set the reward to be 1 when the output answer is correct and -1 otherwise. Then we use MC with $\gamma=1$ to compute the returns. 
We do model selection on a validation dataset sampled from the direct policy model. For GSM8k, we train the value function and ORM for one epoch, while for Game24 and PrOntoQA we train the value function and ORM for 3 epochs.

\textbf{SFT in RLHF alignment}: We utilize GPT2-open-instruct\footnote{https://huggingface.co/vicgalle/gpt2-open-instruct-v1}, a GPT2-Small model supervised-finetuned over several instruction-tuning dataset.

\textbf{Value training in RLHF alignment}: Based on the SFT model, we collect 50 rollouts by the SFT policy for each question in the training set and label their final reward with the reward model. Then we train the value function and ORM for 2 epochs.

Note that here we start training the value function and ORM from the data sampled by the SFT policy model through direct decoding just as an initialization of the value function and ORM. After that \method~can optimize the policy model, value function, and ORM simultaneously by adding new data sampled from tree search into the training buffer.

\textbf{SFT in Chess Endgame}: Currently we use the opensourced behavior-cloning model in \citet{abdulhai2023lmrl} as the SFT model\footnote{https://github.com/abdulhaim/LMRL-Gym}. Given a Forsyth-Edwards Notation (FEN) state description, the language model policy autoregressively outputs SAN(Standard Algebraic Notation)-format actions.

\textbf{Value training in Chess Endgame} Currently we use the opensourced critic value function model trained by Proximal Policy Optimization (PPO) in \citet{abdulhai2023lmrl}. The language model value function receives a FEN state description and outputs a scalar as the value estimation.
\subsection{Details of comparing prompting-based value function and learned value function}
For the prompting-based value function presented in Table \ref{tab:llm_value_function}, we use GPT-3.5-0613 for GPT3.5 model and design few-shot prompts for GPT3.5-policy/value and also LLaMA2-7B. We sample 1 time from GPT3.5 and 3 times from LLaMA2-7B to form the value evaluation.
\subsection{Details of value dataset ablation}
\label{appendix:value_ablation}
Here we introduce the details of building \textit{mixed}, \textit{pure} and \textit{pure,less} datasets on GSM8k for value training in Table~\ref{tab:llm_train_value}. For each model checkpoints 
of 3 epochs during SFT, we first collect 100 outputs per problem in GSM8k training dataset, and then duplicate the overlapped answers, labeled each answer with our training set outcome reward ocracle. 
we sample multiple output sequences with temperature=0.7, top p=1.0 and top k=100.\\
\textbf{\textit{mixed}} dataset: For each deduplicated dataset sampled by models of 3 epochs, we subsample 17 answers per problem.
\\
\textit{\textbf{pure}} dataset: we subsample 50 answers per problem from deduplicated dataset sampled by the last epoch policy model.
\\
\textbf{\textit{pure,less}} dataset: we subsample 17 answers per problem from deduplicated dataset sampled by the last epoch policy model.

For the results in Table~\ref{tab:value_ablation-iterative}, the details of training $\{v, \hat{r}\}_{\phi_{1}}$ can be find in Sec~\ref{appendix:iterative_update-train}. 
We use MC with $\gamma=1$ to compute the returns.
Here we describe the details of training $\{v, \hat{r}\}_{\phi_1}^\text{RL}$, we use the collected 78.7k samples in Sec~\ref{appendix:iterative_update} to optimize $\{v, \hat{r}\}_{\phi_0}$. The training uses a cosine scheduler decaying from lr=2e-5 to 0.0 with a warmup ratio of 0.03, batch size 128 for 3 epochs.

\subsection{Hyperparameter Selection Protocols}
\label{apx:hyper_selection}

\revision{Here we present the selection protocols of hyperparameters.}

\revision{\textbf{Tree search general hyperparameters}: the \textit{tree-max-depth} $d$ limits the search depth of tree and \textit{tree-max-width} $w$ controls the max number of child nodes during node expansion. For the \textit{tree-max-width} $w$, we refer the reader to Appendix \ref{apx: node_expansion} for more discussions. We choose \textit{tree-max-depth} $d$ according to the statistics of the distribution of the number of steps from the dataset sampled by the LLM policy on the training set. Specifically, we statistically analyzed the number distribution of sentences in the training set, and in our experiments, these sentences are split by} `\verb+\n+'. 
\revision{For GSM8K, we set the \textit{tree-max-depth} $d$ at around the 99th percentile of the entire number distribution, to cover most query input and drop the outliers. Game24 has a fixed search depth of 4. For ProntoQA, we set the \textit{tree-max-depth} $d$ at the upper bound of the entire number distribution. For RLHF, this is not a reasoning task with CoT steps, so the depth can be flexible. We set it as the default value. In most cases, the depth of \textit{tree-max-depth} $d$ will not be reached. Because the node expansion will be terminated when we detect our pre-defined stop words in the generation (such as} `\verb+The answer is+' or the `\verb+<EOS>+' token).

\revision{\textbf{Specific hyperparameters for Monte Carlo Tree Search variants}: Basically we adopted the default values from \citet{schrittwieser2020muzero} and \citet{silver2017alphazero} for most of the hyperparameters, such as $c_\text{base}=19652$ in Equation~\ref{equation:puct}, $\tau=1.0$ for MCTS-alpha stochastic search. And for the Dirichlet noise of MCTS-$\alpha$ stochastic search as mentioned in Appendix~\ref{appendix:details-of-aggregation}, we adopted the default value in \citet{silver2017alphazero} as 0.3, which is specified for chess. We do find that in MCTS-$\alpha$, MCTS-Rollout and MCTS, $c_\text{init}$ can affect the balance between exploration and exploitation, and we chose it by running several trials from two possible values: $\{0.3, 3.0\}$. Moreover, for MCTS-$\alpha$, the hyperparameter \textit{num of simulation}, $n_\text{simulation}$ is chosen as 5 for shallow trees (tree max-depths less than of equal to 15 over GSM8k, Game24 and ProntoQA) and 10 in deep trees(a tree max-depth of 64 in RLHF), controlling the search complexity at each step.}

\revision{\textbf{Specific hyperparameters for BFS-/DFS-V}: BFS-V does not have hyperparameters for single search. For DFS-V, the children of a non-leaf node is traversed in a non-decreasing order by its value. For the sake of efficient exploration, we tried 2 heuristics to prune the subtrees, (1) drop the children nodes with lower values by \textit{prune\_ratio}. (2) drop the children nodes lower than a \textit{prune\_value}. The latter is adopted from \citet{yao2023tot}. In our experiments, we tried possible \textit{prune\_value}s from $\{0.5, 0.0, -0.5\}$ or \texttt{None}, we found that setting a high \textit{prune\_value} like $0.5$ or $0.0$ may introduce significant performance drop, however, setting a higher \textit{prune\_value} may introduce very closer answers. Therefore, we finally use \textit{prune\_ratio} for efficient exploration during searching on the tree with DFS-V. We set \textit{prune\_ratio} to be $0.7$ (selected from $\{0.3, 0.5, 0.7\}$) for GSM8k, Game24 and PrOntoQA (tree-max-widths of 6, 6, 20), and 0.95 (selected from $\{0.5, 0.7, 0.9, 0.9\}$)for RLHF alignment task since its much wider(a tree-max-width of 50).}


\subsection{Details of applying each tree search approach}
We present the implementation details and hyperparameters of all tree search approaches here.

Firstly, we refer to Table~\ref{tab:exp-task_setup} for basic settings of each task. We set temperature=1.0, top\_p=1.0, top\_k=100 when using LLM to generate tree actions. To compute logprobs when expand actions in RLHF alignment task trees, we also use a temperature of 1.0.

For MCTS variants including MCTS-$\alpha$, MCTS-Rollout and MCTS, we need to define the hyperparamter in PUCT algorithm:
\begin{equation}
    c_\text{puct} = \log((\sum_b N(s, b) + c_\text{base} + 1) / c_\text{base}) + c_\text{init}
    \label{equation:puct}
\end{equation}

In this paper, we fixed $c_\text{base}$ with 19652 and set $c_\text{init}=3$ for GSM8k, Game24 and RLHF alignment tasks, set $c_\text{init}=0.3$ for PrOntoQA tasks. For Chess Endgame, we manually set $c_\text{puct}=0$.
Specifically, 
in MCTS-$\alpha$, we set the number of simulations before making an action to 5 for GSM8k, Game24 and PrOntoQA, 10 for RLHF alignment and Chess Endgame. We deterministically sampled actions with the largest visit count.
And in MCTS-Rollout, we set an computation upperbound as number of generated tokens or number of model forwards, which is 51200 in GSM8k and Game24, 1800 in PrOntoQA and 5000 in RLHF alignment, 10000 in Chess Endgame.

For DFS-V, the children of a non-leaf node is traversed in a non-decreasing order by value. For efficient exploration, we tried 2 heuristics to prune the subtrees, (1) drop the children nodes with low value by \textit{prune\_ratio}. (2) drop the children nodes lower than a \textit{prune\_value}. We set \textit{prune\_ratio} to be 0.7 for GSM8k, Game24 and PrOntoQA, and 0.95 for RLHF alignment task, 0.6 for Chess Endgame.

All Path@1 results for each tree search approach is conducted with 3 seeds and show the mean and standard deviation. Note that for Path@1 results of tree search approaches, the randomness comes from the node expansion process where we use an LLM to sample candidate actions. While for CoT-SC results, the randomness comes from sampling during direct decoding.

\subsection{Details of aggregation experiments}
\label{appendix:details-of-aggregation}

\revision{Another alternative setting for conducting multiple searches in \textbf{Inter-tree Search}. Inter-tree Search builds a new tree for each new search, which increases the diversity of the search space, with extra computation burdens proportional to the search times. Thus, the intra-setting will have a larger state space compared with intra-tree setting. Our experiment results shown in Fig \ref{fig:2x4_res} (comparing MCTS-$\alpha$ intra-tree and inter-tree settings) also verify the performance gain brought by the larger search space.}

We also present the details of how we sample multiple answers with tree search approaches and aggregate them into a final answer.

\revision{For the results of CoT-SC-Tree on Table~\ref{tab:path1_result} and Table~\ref{tab:ablation_tree_size}, they can be viewed as \textbf{intra-tree} searches. 
For the results in Figure~\ref{fig:2x4_res}, only \textit{MCTS-$\alpha$ inter-trees} were conducted with \textbf{inter-tree} searches, other tree-search algorithms (MCTS, MCTS-Rollout, BFS-V, DFS-V) were all conducted with \textbf{intra-tree} searches}

\revision{Note that for all tree search algorithms except BFS-V, multiple searches are conducted in a sequential manner, while for BFS-V which can actually be regarded as Beam-Search, the number of searches means the number of beam size.}

When sampling multiple intra-tree answers with MCTS-$\alpha$, we use a stochastic sampling setting.
To ensure MCTS-$\alpha$ to explore sufficently, when selecting action of the current node, before doing several times of simulation, we add Dirichlet noise into the language model's prior probability of the current root node $s_0$, i.e. $\pi'_\theta(s_0, a) = (1-\epsilon) \pi_\theta(s_0, a) + \epsilon \eta $, where $\eta \sim \text{Dir}(0.3)$, and we set $\epsilon=0.25$ for both tasks. Actions are sample based on visit count, i.e. $ a \sim \frac{N(s_t, a)^{1 / \tau}}{\sum_b N(s_t, b)^{1 / \tau}}$, where we set $\tau=1$.
After returning with a complete path, we clear the node statistics ($Q(s_t, a_t)$ and $N(s_t, a_t)$) on the tree to eliminate the influence of previous searches, while the tree structure is maintained. This setting is denoted as clear-tree when presented in the table. \textbf{In summary, when we only measure path@1 performance, we adopt MCTS-$\alpha$ (no sampling). But when we measure the aggregation performance, we use MCTS-$\alpha$-intra tree or MCTS-$\alpha$-inter tree. In MCTS-$\alpha$-intra tree we will activate the clear-tree and stochastic sampling setting.}

When sampling multiple answers with other tree-search methods, we only utilize the intra-tree aggregation variant, without stochastic sampling and clear-tree setting. This is because only MCTS-$\alpha$ and MCTS-rollout can conduct the above sampling and clear-tree operation. And we temporarily only apply such setting on MCTS-$\alpha$.

In ORM-vote, since we train the ORM with the reward signal -1 and 1, given a list of $N$ answers to be aggregated, we first normalize its values $\{\hat{r}(y^j)\}_j$ with min-max normalization make them in $[0, 1]$.

\subsection{sampling details of iterative update}
\label{appendix:iterative_update}

We verify the idea of iteratively enhancing language model policy and value function model on the GSM8k and RLHF datasets.

\textbf{Sampling in GSM8k}: When sampling from the 7.5k problems in the GSM8k training dataset, we sample 12 sequences per problem in one sentence-level expanded tree, after deduplication, this results in 78.7k distinct answers, and 73.2\% are correct answers.
The sample parameters are listed in Table~\ref{tab:iterative_update-sample-gsm8k}.


\textbf{Sampling in RLHF alignment}: 
We collect 10 answers for each training set problem sampled by MCTS-$\alpha$. We list the specific hyperparameters in Table~\ref{tab:rlhf_iterative_hyper}.

To collect data for the rejection sampling baseline, we first sample 10 sequences per problem and then use the top 5 sequences for supervised fine-tuning.

\begin{table}[h!]
    \centering
    \caption{Hyperparameters of sampling in GSM8k for LLM decoding(left), tree construction setting(middle), and  MCTS-$\alpha$ setting(right).}
    \label{tab:iterative_update-sample-gsm8k}
    \vspace{-2em}
    \parbox{0.3\linewidth}{
        \begin{tabular}{c  c}
        \toprule
        Hyperparameter &  value \\
        \midrule
        temperature & 1.0 \\
        top\_p & 1.0 \\
        top\_k & 100 \\
        \bottomrule
    \end{tabular}}
    \hfill
    \parbox{0.3\linewidth}{\begin{tabular}{c c}
         \toprule
         Hyperparameter  & value \\
         \midrule
         Tree Max width & 6 \\
         Tree Max depth & 8 \\
         Node & Sentence \\
         \bottomrule
    \end{tabular}}
    \hfill
    \parbox{0.3\linewidth}{\begin{tabular}{c c}
         \toprule
         Hyperparameter  & value \\
         \midrule
         num simulation  & 5\\
         clear tree      & True \\
         stochastic sampling & True \\
         $c_\text{base}$ & 19652 \\
         $c_\text{init}$ & 3 \\
         $\tau$          & 1.0 \\
         \bottomrule
    \end{tabular}}
\end{table}

\begin{table}[h!]
    \centering
    \caption{Hyperparameters of sampling in RLHF alignment for LLM decoding(left), tree construction setting(middle), and  MCTS-$\alpha$ setting(right).}
    \label{tab:rlhf_iterative_hyper}
    \vspace{-2em}
    \parbox{0.3\linewidth}{
        \begin{tabular}{c  c}
        \toprule
        Hyperparameter &  value \\
        \midrule
        temperature & 1.0 \\
        top\_p & 1.0 \\
        top\_k & 50 \\
        \bottomrule
    \end{tabular}}
    \hfill
    \parbox{0.3\linewidth}{\begin{tabular}{c c}
         \toprule
         Hyperparameter  & value \\
         \midrule
         Tree Max width & 50 \\
         Tree Max depth & 64 \\
         Node & Token \\
         \bottomrule
    \end{tabular}}
    \hfill
    \parbox{0.3\linewidth}{\begin{tabular}{c c}
         \toprule
         Hyperparameter  & value \\
         \midrule
         num simulation  & 5 \\
         clear tree      & True \\
         stochastic sampling & True \\
         $c_\text{base}$ & 19652 \\
         $c_\text{init}$ & 3 \\
         $\tau$          & 1.0 \\
         \bottomrule
    \end{tabular}}
\end{table}

\subsection{Training details of iterative update}
\label{appendix:iterative_update-train}
\textbf{Policy training in GSM8k}:
We construct the dataset for supervised finetuning by combining data in the training dataset with 57.6k correct answers sampled in Sec~\ref{appendix:iterative_update} which results in 64.1k distinct correct answers.
And we train the new policy model $\pi_{\theta_1}$ from the starting base model LLaMA2-7b for 3 epochs, following \citet{yuan2023rft}. The training setting is the same as described in Sec~\ref{appendix:sft_value_training}. 

\textbf{Value training in GSM8k}: We construct the dataset for value and ORM training by combining the data used to train $\{v, \hat{r}\}_{\phi_0}$ with 78.7k answers sampled by MCTS-$\alpha$ in Sec~\ref{appendix:iterative_update}. To fairly compare $\{v, \hat{r}\}_{\phi_1}$ with $\{v, \hat{r}\}_{\phi_0}$, we drop samples in the former dataset to keep at most $51-12=39$ answers per problem resulting in 359k distinct answers.
And we train the new value function $\{v, \hat{r}\}_{\phi_1}$ from the value model with its initial weight(before being updated on any data) for 3 epochs. The training setting is the same as described in Sec~\ref{appendix:sft_value_training}.

\textbf{Policy training in RLHF alignment}: For the MCTS-$\alpha$'s training, we subsample the top 5 answers from the full 10 candidates (mentioned in Appendix~\ref{appendix:iterative_update}) to serve as the SFT dataset. For the RFT n=5 baseline, we subsample the top 5 answers from 50 direct decodings as the SFT dataset. For the training of the PPO algorithm, we adopt the implementation from trlx\footnote{https://github.com/CarperAI/trlx}. We sample 20 answers for each question in total, which maintains the same level of token consumption during the PPO rollouts as that of MCTS-$\alpha$. 

\textbf{Value training in RLHF alignment}: We construct the value and ORM dataset by mixing data from SFT-policy direct decoding and from MCTS-$\alpha$. To make the comparison fair, the new value function's training utilizes the same amount of data as the old one by subsampling 40 answers (from 50 shown in Appendix~\ref{appendix:sft_value_training}) from direct decoding data and all 10 answers (shown in Appendix~\ref{appendix:iterative_update}) generated by MCTS-$\alpha$. We train our value function with learning rate 2e-5 and cosine scheduler from the initial model (instead of continuing training from the old value function) for 2 epochs.

\section{Qualitative Results}
Here we present some qualitative results of generation for different methods in Table~\ref{tab:qr_game24} (for Game24), \revision{Table~\ref{tab:qr_gsm8k} (for GSM8k), Table~\ref{tab:qr_prontoqa} (for ProntoQA)} and \ref{tab:qr_rlhf} (for the RLHF alignment task).
\label{appendix:qualitative_results}

\begin{table}[htbp]
    \centering
    \begin{tabular}{c p{55mm} c}
        \toprule
        \midrule
        \multicolumn{3}{l}{Problem: 4 8 9 13} \\
        \midrule[\heavyrulewidth]
         \shortstack{Methods} & Response & \shortstack{Correctness} \\ 
         \midrule
         CoT & 8 + 13 = 21 (left: 4 9 21)
         
               4 - 9 = -5 (left: -5 21)  
               
               21 - -5 = 26 (left: 26) 
               
               The answer is (8 + 13) - (4 - 9) = 26 & No  \\
         \midrule
         BFS-V & 9 + 3 = 12 (left: 4 8 12)
         
               12 / 4 = 3 (left: 3 8)
               
               3 * 8 = 24 (left: 24)
               
               The answer is (9 + 3) / 4 * 8 = 24 & No \\
         \midrule
        MCTS-$\alpha$ & 13 - 9 = 4 (left: 4 4 8) 
        
        4 * 4 = 16 (left: 8 16)     
                       
        16 + 8 = 24 (left: 24)
        
        The answer is 4 * (13 - 9) + 8 = 24 & Yes \\
        \midrule
        MCTS-Rollout & 13 - 9 = 4 (left: 4 4 8) 
        
                      4 * 4 = 16 (left: 8 16)
                      
                      16 + 8 = 24 (left: 24)
                      
                      The answer is (13 - 9) * 4 + 8 = 24 & Yes  \\
        \bottomrule
    \end{tabular}
    \caption{Qualitative examples of Game24 outputs}
    \label{tab:qr_game24}
\end{table}

\begin{table}[htbp]
    \centering
    \begin{tabular}{c p{90mm} c}
        \toprule
        \midrule
        \multicolumn{3}{p{\dimexpr\textwidth-2\tabcolsep\relax}}{Question: A new program had 60 downloads in the first month. The number of downloads in the second month was three times as many as the downloads in the first month, but then reduced by 30\% in the third month. How many downloads did the program have total over the three months?} \\
        \midrule[\heavyrulewidth]
         \shortstack{Methods} & Response & \shortstack{Correct} \\ 
         \midrule
         CoT & The number of downloads in the second month was 3*60 = 180
         
The number of downloads in the third month was 180*0.30 = 54

The total number of downloads for the three months was 60+180+54 = 294

The answer is 294
                & No  \\
         \midrule
         BFS-V & In the second month, the program had 60*3=180 downloads
         
In the third month, the program had 180*30/100=54 downloads which was 30\% less than the previous month

Total downloads for the program in the three months was 60+180+54=294

The answer is 294
                & No \\
         \midrule
        MCTS-$\alpha$ & The program had 60 x 3 = 180 downloads in the second month.
        
The downloads in the third month was 180 x 0.30 = 54 downloads lesser than the previous month.

So, the program had 180 - 54 = 126 downloads in the third month.

Therefore, the program had 60 + 180 + 126 = 366 downloads over the three months.

The answer is 366
                    &  Yes \\
        \midrule
        MCTS-Rollout & The number of downloads in the second month was 3 times as many as the downloads in the first month, which means the program had 3*60 = 180 downloads in the second month.
        
The total number of downloads for the two months was 180+60 = 240
In the third month, the program had 180-30/100*180 = 120 downloads, 30\% of the number of downloads it had in the second month.

The total number of downloads for the three months was 240+120 = 360

The answer is 360
                    &  No \\
        \bottomrule
    \end{tabular}
    \caption{Qualitative examples of GSM8k outputs}
    \label{tab:qr_gsm8k}
\end{table}

\begin{table}[htbp]
    \centering
    \begin{tabular}{c p{40mm} c}
        \toprule
        \midrule
        \multicolumn{3}{p{\dimexpr\textwidth-19\tabcolsep\relax}}{Question: Butterflies are lepidopterans. Every arthropod is small. Whales are not small. Invertebrates are animals. Every insect is an arthropod. Lepidopterans are insects. Every insect is six-legged. Every arthropod is an invertebrate. Animals are multicellular. Polly is a lepidopteran. Is the statement "Polly is not small" true or false?} \\
        \midrule[\heavyrulewidth]
         \shortstack{Methods} & Response & \shortstack{Correct} \\ 
         \midrule
         CoT & Polly is a lepidopteran.
         
                Lepidopterans are insects.
                
                Polly is an insect.
                
                Every insect is an arthropod.
                
                Polly is an arthropod.
                
                Every arthropod is small.
                
                Polly is small.
                
                The answer is true. 
                & No  \\
         \midrule
         BFS-V & Polly is a lepidopteran.
         
                Lepidopterans are insects.
                
                Polly is an insect.
                
                Every insect is an arthropod.
                
                Polly is an arthropod.
                
                Every arthropod is small.
                
                Polly is small.
                
                Polly is small.
                
                The answer is false. 
                & Yes \\
         \midrule
        MCTS-$\alpha$ & Polly is a lepidopteran.
        
                    Lepidopterans are insects.
                    
                    Polly is an insect.
                    
                    Every insect is an arthropod.
                    
                    Polly is an arthropod.
                    
                    Every arthropod is small.
                    
                    Polly is small.
                    
                    The answer is false.
                    &  Yes \\
        \midrule
        MCTS-Rollout & Polly is a lepidopteran.
        
        Lepidopterans are insects.
        
        Polly is an insect.
        
        Every insect is an arthropod.
        
        Polly is an arthropod.
        
        Every arthropod is small.
        
        Polly is small.
        
        The answer is false.
                    &  Yes \\
        \bottomrule
    \end{tabular}
    \caption{Qualitative examples of ProntoQA outputs}
    \label{tab:qr_prontoqa}
\end{table}

         
\begin{table}[htbp]
    \centering
    \begin{tabular}{c p{55mm} c}
        \toprule
        \midrule
        \multicolumn{3}{l}{Prompt: What careers am I good at.} \\
        \midrule[\heavyrulewidth]
         \shortstack{Methods} & Response & \shortstack{Reward} \\ 
         \midrule
         CoT & You can be a great career at any career you choose. Depending on your interests and skills, you may be able to find a variety of jobs that align with your interests and skills. & -2.625  \\
         \midrule
         BFS-V & Some careers great careers to do professionally are entrepreneur in healthcare policy policy advocacy consulting public diplomacy leadership policy innovation in global issues managing information theory. Some great careers to achieve excellence in healthcare management include doctor psychologist & -2.03 \\
         \midrule
        MCTS-$\alpha$ & You may be good at a variety of careers, depending on your interests and skills. Some of the most popular careers include software engineer, data scientist. You may also be good at a variety of fields such as finance, marketing, and business administration. &  -1.67 \\
        \bottomrule
    \end{tabular}
    \caption{Qualitative examples of RLHF outputs}
    \label{tab:qr_rlhf}
\end{table}

\section{Detailed Experiment Results}
\subsection{Results per task per aggregation}
We show detailed results of GSM8k on Table~\ref{tab:detail-results-gsm8k}, results of Game24 on Table~\ref{tab:detail-results-game24}, Table~\ref{tab:detail-results-prontoqa} for PrOnToQA and Table~\ref{tab:detail-results-rlhf} for RLHF alignment. \revision{Due to the limit of computation resources, we show the results under 1 seed except for the path@1 results.}
\begin{table}[!h]
    \centering
    \caption{Detailed Results in GSM8k}
    \label{tab:detail-results-gsm8k}
    \begin{tabular}{c | ccccccccc}
        \toprule
        Method       & N   & Majority-vote & ORM-vote & ORM-max & \#Token  \\
        \midrule
        CoT          & -   & 41.4  & 41.4 & 41.4    & 0.1k\\
        \midrule
        CoT-SC       & 1   & 38.21 & 38.21 & 38.21  & 0.1k \\ 
        CoT-SC       & 10  & 51.93 & 57.47 & 53.83  & 1k \\ 
        CoT-SC       & 20  & 54.44 & 59.44 & 54.74  & 2k \\ 
        CoT-SC       & 50  & 56.79 & 61.03 & 54.44  & 5k \\ 
        CoT-SC       & 100 & 58.15 & 62.70 & 53.68  & 10k \\ 
        \midrule
        CoT-SC-Tree & 1 & 37.91 & 37.91 & 37.91 & 0.1k \\
        CoT-SC-Tree & 10 & 50.19 & 53.15 & 50.95 & 0.8k \\
        CoT-SC-Tree & 20 & 52.69 & 55.12 & 52.84 & 1.3k \\
        CoT-SC-Tree & 50 & 54.51 & 57.16 & 53.45 & 2.7k \\
        \midrule
        BFS-V & 1 & 52.5 & 52.5 & 52.5 & 0.5k \\
        BFS-V & 10 & 58.98 & 56.25 & 54.97 & 3.1k \\
        BFS-V & 20 & 58.91 & 56.79 & 52.39 & 5.3k \\
        BFS-V & 50 & 59.29 & 59.36 & 53.22 & 10.1k \\
        \midrule
        DFS-V & 1 & 51.8 & 51.8 & 51.8 & 0.5k \\
        DFS-V & 10 & 57.09 & 56.18 & 54.89 & 1.2k \\
        DFS-V & 20 & 58.23 & 58.38 & 55.19 & 1.6k \\
        DFS-V & 50 & 58.98 & 58.98 & 55.35 & 2.1k \\
        \midrule
        MCTS-$\alpha$ (no sampling) & 1 & 51.9 & 51.9 & 51.9  & 0.5k \\
        
        \midrule
        MCTS-$\alpha$-intra tree & 1 & 46.78 & 46.78 & 46.78  & 0.7k \\
        MCTS-$\alpha$-intra tree & 10 & 57.85 & 56.86 & 54.36 & 3.4k \\
        MCTS-$\alpha$-intra tree & 20 & 58.83 & 58.23 & 55.19 & 5.3k \\
        \midrule
        MCTS-$\alpha$-inter trees & 1 & 51.9 & 51.9 & 51.9  & 0.5k \\ 
        MCTS-$\alpha$-inter trees & 10 & 57.92 & 58.53 & 55.34 & 5.5k \\ 
        MCTS-$\alpha$-inter trees & 20 & 58.83 & 59.06 & 54.97 & 11.1k \\ 
        MCTS-$\alpha$-inter trees & 50 & 58.76 & 61.26 & 53.98 & 27.8k \\ 
        \midrule
        MCTS & 1 & 52.2 & 52.2 & 52.2 & 0.5k \\ 
        MCTS & 10 & 57.92 & 55.72 & 53.75 & 2.4k \\
        MCTS & 20 & 58.61 & 56.79 & 54.74 & 4.0k \\
        MCTS & 50 & 59.36 & 58.23 & 53.75 & 7.5k \\
        \midrule
        MCTS-Rollout & 1 & 47.8 & 47.8 & 47.8 & 3.4k \\
        MCTS-Rollout & 10 & 51.10 & 50.49 & 49.81, & 5.4k \\
        MCTS-Rollout & 20 & 51.86 & 51.25 & 50.19 & 6.1k \\
        MCTS-Rollout & 50 & 52.69 & 52.24 & 50.49 & 7.2k \\
        \bottomrule
    \end{tabular}
\end{table}

\begin{table}[!h]
    \centering
    \caption{Detailed Results in Game24}
    \label{tab:detail-results-game24}
    \begin{tabular}{cc | cccccccc}
        \toprule
        Method       & N   & Majority-vote & ORM-vote & ORM-max & \#Token  \\
        \midrule
        CoT &  - & 12.7 & 12.7 & 12.7 & 0.1k \\
        \midrule
        CoT-SC & 1 & 9.94 & 9.94 & 9.94 & 0.1k \\ 
        CoT-SC & 10 & 13.54 & 50.83 & 50.83 & 0.8k \\
        CoT-SC & 20 & 14.36 & 65.75 & 65.47 & 1.6k \\
        CoT-SC & 50 & 16.30 & 78.45 & 78.45 & 4.0k \\
        CoT-SC & 100 & 18.23 & 84.25 & 84.53 & 7.9k \\
        \midrule
        CoT-SC-Tree & 1 & 9.67 & 9.67 & 9.67 & 0.1k \\
        CoT-SC-Tree & 10 & 11.33 & 48.34 & 48.34 & 0.7k \\
        CoT-SC-Tree & 20 & 13.26 & 61.60 & 62.15 & 1.1k \\
        CoT-SC-Tree & 50 & 16.57 & 69.61 & 69.89 & 2.0k \\
        \midrule
        BFS-V & 1 & 64.8 & 64.8 & 64.8 & 0.4k \\
        BFS-V & 10 & 47.79 & 70.72 & 70.99 & 1.6k \\
        BFS-V & 20 & 27.62 & 69.34 & 69.34 & 2.3k \\
        BFS-V & 50 & 7.18 & 70.17 & 70.72 & 3.7k \\
        \midrule
        DFS-V & 1 & 66.3 & 66.3 & 66.3 & 0.4k \\
        DFS-V & 10 & 55.25 & 69.06 & 69.34 & 0.9k \\
        DFS-V & 20 & 54.14 & 69.34 & 69.61 & 1.0k \\
        \midrule
        MCTS-$\alpha$ (no sampling) & 1 & 63.3 & 63.3 & 63.3 & 0.4k \\
        \midrule
        MCTS-$\alpha$-intra tree & 1 & 64.36 & 64.36 & 64.36 & 0.4k \\
        MCTS-$\alpha$-intra tree & 10 & 66.85 & 67.68 & 63.90 & 0.9k \\
        MCTS-$\alpha$-intra tree & 20 & 67.13 & 69.34 & 68.78 & 1.1k \\
        MCTS-$\alpha$-intra tree & 50 & 67.96 & 69.89 & 69.34 & 1.4k \\
        \midrule
        MCTS-$\alpha$-inter trees & 1 & 63.3 & 63.3 & 63.3 & 0.4k \\ 
        MCTS-$\alpha$-inter trees & 10 & 72.65 & 82.87 & 82.32 & 4.1k \\
        MCTS-$\alpha$-inter trees & 20 & 72.93 & 84.25 & 83.15 & 8.3k \\
        \midrule
        MCTS & 1 & 64.0 & 64.0 & 64.0 & 0.4k \\ 
        MCTS & 10 & 70.44 & 70.72 & 70.17 & 0.8k \\ 
        MCTS & 20 & 72.10 & 72.10 & 71.27 & 1.1k \\
        MCTS & 50 & 72.38 & 72.38 & 71.55 & 1.6k \\
        \midrule
        MCTS-Rollout & 1 & 71.3 & 71.3 & 71.3 &  0.7k\\ 
        MCTS-Rollout & 10 & 73.48 & 73.20 & 72.65 & 0.9k \\
        MCTS-Rollout & 20 & 73.48 & 73.48 & 72.38 & 1.0k \\
        MCTS-Rollout & 50 & 73.48 & 73.48 & 72.38 & 1.1k \\ 
        \bottomrule
    \end{tabular}
\end{table}

\begin{table}[!ht]
    \centering
    \caption{Detailed Results in PrOntoQA}
    \label{tab:detail-results-prontoqa}
    \begin{tabular}{cc | cccccccc}
        \toprule
        Method       & N   & Majority-vote & ORM-vote & ORM-max & \#Token  \\
        \midrule
        CoT    & - & 48.8 & 48.8 & 48.8 & 92 \\
        \midrule
        CoT-SC & 1 & 54.40 & 54.40 & 54.40 & 91.25 \\
        CoT-SC & 3 & 63.60 & 82.40 & 82.40 & 273.75 \\
        CoT-SC & 10 & 58.40 & 97.80 & 97.80 & 912.55 \\
        CoT-SC & 20 & 57.00 & 99.80 & 99.80 & 1.8k \\
        \midrule
        CoT-SC-Tree & 1 & 50.20 & 50.20 & 50.20 & 82.02 \\
        CoT-SC-Tree & 10 & 62.40 & 98.40 & 98.40 & 413.58 \\ 
        CoT-SC-Tree & 20 & 61.00 & 99.40 & 99.40 & 632.91 \\ 
        \midrule
        BFS-V & 1 & 94.40 & 94.40 & 94.40 & 125.52 \\ 
        BFS-V & 10 & 99.00 & 100.00 & 100.00 & 837.78 \\
        BFS-V & 20 & 98.60 & 99.80 & 99.80 & 1.5k \\ 
        \midrule
        DFS-V & 1 & 93.30 & 93.30 & 93.30 & 124.46 \\
        DFS-V & 10 & 95.60 & 96.40 & 96.40 & 187.59 \\
        DFS-V & 20 & 95.60 & 96.40 & 96.40 & 193.91 \\
        \midrule
        MCTS-$\alpha$ (no sampling) & 1 & 99.40 & 99.40 & 99.40 & 183.66 \\ 
        \midrule
        MCTS-$\alpha$-intra tree & 1 & 97.20 & 97.20 & 97.20 & 208.68 \\ 
        MCTS-$\alpha$-intra tree & 10 & 99.80 & 99.80 & 99.80 & 364.96 \\
        MCTS-$\alpha$-intra tree & 20 & 99.80 & 99.80 & 99.80 & 441.31 \\
        \midrule
        MCTS-$\alpha$-inter trees & 1 & 99.40 & 99.40 & 99.40 & 183.66 \\ 
        MCTS-$\alpha$-inter trees & 10 & 100.00 & 100.00 & 100.00 & 1.9k \\ 
        MCTS-$\alpha$-inter trees & 20 & 100.00 & 100.00 & 100.00 & 3.8k \\ 
        \midrule
        MCTS & 1 & 94.20 & 94.20 & 94.20 & 126.65 \\ 
        MCTS & 10 & 99.60 & 99.60 & 99.60 & 182.88 \\
        MCTS & 20 & 100.00 & 100.00 & 100.00 & 240.16 \\
        \midrule
        MCTS-Rollout & 1 & 96.90 & 96.90 & 96.90 & 210.41 \\ 
        MCTS-Rollout & 10 & 99.20 & 99.20 & 99.20 & 220.16 \\ 
        MCTS-Rollout & 20 & 99.20 & 99.20 & 99.20 & 224.16 \\ 
        \bottomrule
    \end{tabular}
\end{table}

\begin{table}[!ht]
    \centering
    \caption{Detailed Results in RLHF alignment}
    \label{tab:detail-results-rlhf}
    \begin{tabular}{cc | ccc}
         \toprule
        Method       & N   & Mean & Best & \#Forward  \\
        \midrule
        CoT & 1 & 0.387 & 0.387 & 57.8 \\
        \midrule
        CoT-SC & 1 & -0.164 & -0.164 & 58 \\ 
        CoT-SC & 10 & -0.182 & 1.592 & 0.6k \\ 
        CoT-SC & 20 & -0.175 & 1.972 & 1.2k \\ 
        CoT-SC & 50 & -0.176 & 2.411 & 2.9k \\ 
        \midrule
        BFS-V & 1 & -1.295 & -1.295 & 61.8 \\ 
        BFS-V & 10 & -1.523 & -1.065 & 0.6k \\ 
        BFS-V & 20 & -1.520 & -0.948 & 1.2k \\ 
        BFS-V & 50 & -1.474 & -0.813 & 3.1k \\ 
        \midrule
        DFS-V & 1 & -1.295 & -1.295 & 61.8 \\ 
        DFS-V & 10 & -1.498 & -1.067 & 67.8 \\ 
        DFS-V & 20 & -1.507 & -0.985 & 71.8 \\ 
        DFS-V & 50 & -1.503 & -0.86 & 85.8 \\
        \midrule
        MCTS-$\alpha$ (no sampling) & 1 & 2.221 & 2.221 & 186 \\ 
        \midrule
        MCTS-$\alpha$-intra tree & 1 & 1.538 & 1.538 & 198.50 \\ 
        MCTS-$\alpha$-intra tree & 10 & 1.527 & 3.052 & 1.6k \\
        MCTS-$\alpha$-intra tree & 20 & 1.533 & 3.311 & 3.1k \\ 
        \midrule 
        MCTS & 1 & -1.295 & -1.295 & 61.8 \\ 
        MCTS & 10 & -1.146 & 0.160 & 0.6k \\ 
        MCTS & 20 & -1.08 & 0.528 & 1.2k \\ 
        MCTS & 50 & -0.961 & 0.981 & 2.8k \\ 
        \midrule
        MCTS-Rollout & 1 & 1.925 & 1.925 & 0.8k \\ 
        MCTS-Rollout & 10 & 2.278 & 2.540 & 1.1k \\ 
        MCTS-Rollout & 20 & 2.376 & 2.643 & 1.2k \\ 
        MCTS-Rollout & 50 & 2.491 & 2.746 & 1.3k \\ 
        \bottomrule
    \end{tabular}
\end{table}